\documentclass{article}

\PassOptionsToPackage{}{natbib}

\usepackage{iclr2025_conference,times}
\iclrfinalcopy

\usepackage{amsmath}
\usepackage{amssymb}
\usepackage{mathtools}
\usepackage{amsthm}
\usepackage{comment}
\usepackage{url}   
\usepackage{algpseudocode} 
\usepackage{amsfonts}       %
\usepackage{nicefrac}       %
\usepackage{microtype}      %
\usepackage{color}
\usepackage{graphicx}
\usepackage{tcolorbox}
\usepackage{tikz}
\usepackage{natbib}
\usepackage{subfigure}

\usepackage{caption}
\usepackage{multirow}
\usepackage{wrapfig}
\usepackage{enumitem}

\usepackage[backref=page, colorlinks=True, linkcolor=teal, citecolor=teal]{hyperref}
\usepackage{tabularx}

\usepackage{cleveref}

\usepackage{varwidth}
\usepackage{lipsum}
\usepackage{multicol}
\newsavebox{\algbox}
\usepackage{import}
\usetikzlibrary{bayesnet}
\usetikzlibrary{arrows}
\usepackage{makecell}

\usepackage[utf8]{inputenc} 
\usepackage[T1]{fontenc}    
\usepackage{soul}
\usepackage{bm}
\usepackage{listings}
\usepackage[title]{appendix}
\usepackage[bottom]{footmisc}
\usepackage[makeroom]{cancel}
\usepackage[normalem]{ulem}
\setcitestyle{authoryear,open={(},close={)}}
\usepackage{array}
\newcolumntype{P}[1]{>{\centering\arraybackslash}p{#1}}
\usepackage{booktabs}
\usepackage{xcolor}         
\usepackage{bm}

\usepackage{pifont}

\usepackage{subcaption}

\def\x{\mathbf{x}}
\def\z{\mathbf{z}}
\def\y{\mathbf{y}}

\def\s{\mathbf{s}}

\def\X{\mathcal{X}}

\def\R{\mathbb{R}}

\DeclareMathOperator*{\argmax}{arg\,max}
\DeclareMathOperator*{\argmin}{arg\,min}

\newcommand{\yv}[1]{\textcolor{olive}{[YV: #1]}}
\usepackage{algorithm}
\usepackage{algpseudocode}

\newtheorem{definition}{Definition}
\newtheorem{proposition}{Proposition}

\newcommand{\given }{ \, | \, }
\newcommand{\ourmethod }{\text{RISE }}

\title{Robust Simulation-Based Inference\\ under Missing Data via Neural Processes}

\author{Yogesh Verma, Ayush Bharti\\
Department of Computer Science, Aalto University \\
\texttt{\{yogesh.verma, ayush.bharti\}@aalto.fi}\\
\And 
Vikas Garg\\
YaiYai Ltd and Aalto University\\
\texttt{vgarg@csail.mit.edu}
}

\iclrfinalcopy 

\begin{document}
\maketitle

\begin{abstract}

Simulation-based inference (SBI) methods typically require fully observed data to infer parameters of models with intractable likelihood functions. However, datasets often contain missing values due to incomplete observations, data corruptions (common in astrophysics), or instrument limitations (e.g., in high-energy physics applications). In such scenarios, missing data must be imputed before applying any SBI method. We formalize the problem of missing data in SBI and demonstrate that naive imputation methods can introduce bias in the estimation of SBI posterior. We also introduce a novel amortized method that addresses this issue by jointly learning the imputation model and the inference network within a neural posterior estimation (NPE) framework. Extensive empirical results on SBI benchmarks show that our approach provides robust inference outcomes compared to standard baselines for varying levels of missing data. Moreover, we demonstrate the merits of our imputation model on two real-world bioactivity datasets (Adrenergic and Kinase assays).  Code is available at \url{https://github.com/Aalto-QuML/RISE}.


    
\end{abstract}


\section{Introduction}

Mechanistic models for studying complex physical or biological phenomena have become indispensable tools in research fields as diverse as genetics \citep{Riesselman2018}, epidemiology \citep{Kypraios2017}, gravitational wave astronomy \citep{Dax2021}, and radio propagation \citep{Bharti2022}. However, fitting such models to observational data can be challenging due to the intractability of their likelihood functions, which renders standard Bayesian inference methods inapplicable. Simulation-based inference (SBI) methods \citep{cranmer2020frontier} tackle this issue by relying on forward simulations from the model instead of evaluating the likelihood. These simulations are then either used to train a conditional density estimator \citep{papamakarios2016fast, Lueckmann2017, greenberg2019automatic, papamakarios2019sequential, Radev2022}, or to measure distance with the observed data \citep{Sisson2018, Briol2019MMD, pesonen2023abc}, to approximately estimate the posterior distribution of the model parameters of interest. 

SBI methods implicitly assume that the observed data distribution belongs to the family of distributions induced by the model; i.e., the model is \emph{well-specified}. However, this assumption is often violated in practice where models tend to be \emph{misspecified} since the complex real-world phenomena under study are not accurately represented.  
Even if the model is well-specified, the data collection mechanism might hinder the applicability of SBI methods since it can induce missing data due to, for instance, incomplete observations \citep{luken2021missing}, instrument limitations \citep{kasak2024machine}, or unfavorable experimental conditions.

\begin{wrapfigure}[24]{r}{0.45\textwidth}
    \vspace{-7.2pt}
    \includegraphics[width=0.42\textwidth]{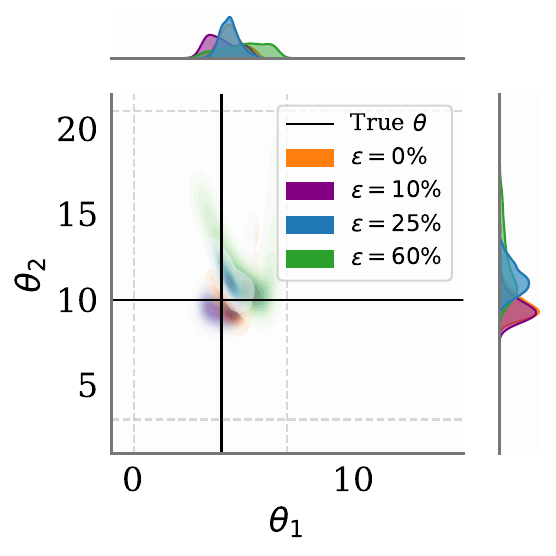}
     \caption{\textbf{Effect of missing data on SBI.} NPE posterior for the two-parameter Ricker model \citep{Wood2010} where the method of \citet{wang2024missing} (with zero augmentation) is used to handle $\varepsilon\%$ of values missing in the data. As $\varepsilon$ increases, the NPE posteriors become  biased and drift away from the true parameter value, denoted by the black lines.}
     \label{fig:miss_data}
\end{wrapfigure}

Although the former problem of model misspecification has been studied in a number of works \citep{Frazier2020ABC, Dellaporta2022, 2021Fujisawa, Bharti22ICML, Ward2022robust, schmitt2023detecting, gloeckler2023adversarial, huang2024learning, gao2024generalized, kelly2024misspecificationrobust}, the latter problem of missing data in SBI has received relatively less attention. A notable exception is the work of \citet{wang2024missing}, which attempts to handle missing data by augmenting and imputing constant values (e.g., zero or sample mean) and performing inference with a binary mask indicator. However, this approach can lead to biased estimates, reduced variability, and distorted relationships between variables \citep{graham2007review}. This is exemplified in \Cref{fig:miss_data} where we investigate the impact of missing data on neural posterior estimation (NPE, \citet{papamakarios2016fast})---a popular SBI method---on a population genetics model. We observe that simply incorporating missing values and their corresponding masks in NPE methods as in \citet{wang2024missing} leads to biased posterior estimates.
Other SBI works that address missing data include \cite{lueckmann2017flexible} and  \cite{gloeckler2024all}, however, they fail to account for the underlying mechanism that leads to missing values in the data.

Outside of SBI, the problem of missing data has been extensively studied \citep{van2011mice}, with \citet{rubin1976inference} categorizing it into three types: missing completely at random (MCAR), missing at random (MAR), and missing not at random (MNAR). Recent advances in machine learning have led to the development of novel methods for addressing this problem using generative adversarial networks (GANs, \citet{luo2018multivariate, yoon2018gain, li2019misgan, yoon2020gamin}), variational autoencoders (VAEs, \citet{ nazabal2020handling, collier2020vaes,mattei2019miwae,ipsen2020not,ghalebikesabi2021deepgenerativepatternsetmixture}), Gaussian processes \citep{casale2018gaussian,fortuin2020gp,ramchandran2021longitudinal,ong2024learning}, and optimal transport \citep{muzellec2020missingdataimputationusing, zhao2023transformed,  vo2024optimal}. These methods offer new perspectives on the problem of missing data imputation, but their application has been primarily limited to predicting missing values. Notably, they have not been developed for inference\\ over missing values, which remains a significant challenge for SBI. 

\vspace{-1ex}

\paragraph{Contributions.} In this paper, we introduce a novel SBI method that is robust to shift in the posterior distribution due to missing data. Our method, named RISE (short for ``Robust Inference under imputed SimulatEd data''), jointly performs imputation and inference by combining NPE with latent neural processes \citep{foong2020meta}.  Doing so allows us to learn an \emph{amortized} model unlike other robust SBI methods in the literature, and to handle missing data under different assumptions \citep{little2019statistical}.
We summarize our main contributions below:
\vspace{-1ex}
 \begin{itemize}
     \item we motivate the problem of missing data in SBI, arguing how it can induce bias in posterior estimation;
     \item we propose RISE, an {amortized} method, that jointly learns an imputation and inference network to deal with missing data; 
     \item RISE outperforms competing baselines in inference and imputation tasks across varying levels of missingness, demonstrating robust performance in settings entailing missing data.
 \end{itemize}


\section{Preliminaries}

Consider a simulator-based model $p(\cdot \given \theta )$ that takes in a parameter vector $\theta \in \Theta \subseteq \mathbb{R}^p$ and maps it to a point $\x = [x_1, \dots, x_d]^\top$ in some data space $\X \subseteq \R^d$. We assume that $p(\cdot \given \theta )$ is intractable, meaning that its associated likelihood function is unavailable and cannot be evaluated point-wise. However, in our setting, generating independent and identically distributed (iid) realisations $\x \sim p(\cdot \given \theta )$ for a fixed $\theta$ is straightforward.
Given a dataset $\tilde \x$ collected via real-world experiments from some true data-generating process and a prior distribution on the parameters $p(\theta)$, we are interested in approximating the posterior distribution $p(\theta \given \tilde \x) \propto p(\tilde \x \given \theta) p(\theta)$. This can be achieved, for instance, using the popular neural posterior estimation (NPE) method, which we now introduce.  

\paragraph{Neural posterior estimation.} NPE \citep{papamakarios2016fast} involves training conditional density estimators, such as normalizing flows \citep{papamakarios2021normalizing}, to learn a mapping from each datum $\x$ to the posterior distribution $p(\theta \given \x)$. 
Specifically, we can approximate the posterior distribution with $q_\phi(\theta \given \x)$ using learnable parameters $\phi$.  In particular, we can train $q_\phi$ by minimizing an empirical loss 
\begin{equation}\label{eq:npe_loss}
    \ell_{\mathrm{NPE}}(\phi) \triangleq -\frac{1}{n} \sum_{i=1}^n \log q_{\phi} (\theta_i \given \x_i) \approx - \mathbb{E}_{\theta \sim p(\theta)}[\mathbb{E}_{\x \sim p(\x \given \theta)}[\log q_{\phi} (\theta \given \x)]],
\end{equation}
using the dataset $\{(\theta_i, \x_i)\}_{i=1}^n$ simulated from the joint distribution $p(\theta, \x) = p(\x \given \theta) p(\theta)$. When the data space $\mathcal{X}$ is high-dimensional, or there are multiple observations $\mathbf{X} = [\mathbf{x}^{(1)}, \dots, \mathbf{x}^{(m)}]$ for each $\theta$, we can use a summary function  $\eta:\mathcal{X}\rightarrow \mathcal{S}$ (such as a deep set \citep{zaheer2017deep}) to enable a condensed representation. Assuming that the summary function is parameterized by $\kappa$, the joint NPE loss with respect to both $\phi$ and $\kappa$ can be defined as $\ell_{\mathrm{NPE}}(\phi, \kappa) \triangleq -\frac{1}{n} \sum_{i=1}^n \log q_{\phi} (\theta_i \given \eta_\kappa(\x_i))$.
Once both $q_\phi$ and $\eta_\kappa$ are trained, the NPE posterior estimate $q_{\hat{\phi}}(\theta \given \eta_{\hat \kappa}(\tilde{\x}))$ for any given real data  $\tilde{\x}$ is obtained by a simple forward pass of $\tilde{\x}$ through the trained networks, making NPEs amortized. We now provide a brief background on the missing data problem, which is the focus of this work.

\paragraph{Missing data background.} In the context of missing data, each data sample is composed of an observed part $\mathbf{x}_{\text{obs}}$ and a missing (or unobserved) part $\mathbf{x}_{\text{mis}}$ such that $\x = (\mathbf{x}_{\text{obs}}, \mathbf{x}_{\text{mis}})$. The missingness pattern for each $\x$ is described by a binary mask variable $\mathbf{s} \in \{0,1\}^d$, where $s_i = 1$ if the element $x_i$ is observed and $s_i = 0$ if $x_i$ is missing, $i \in \{1, \dots, d\}$. The joint distribution of $\x$ and $\mathbf{s}$ can be factorized as $p(\x, \mathbf{s}) = p(\mathbf{s} \given \x)p(\x)$. Based on specific assumptions about what the conditional distribution of the mask (or the missingness mechanism) depends on, three different scenarios arise  \citep{little2019statistical}: (i) missing-completely-at-random (MCAR), where $p(\mathbf{s} \given \x) = p(\mathbf{s})$; (ii) missing-at-random (MAR), where $p(\mathbf{s} \given \x) = p(\mathbf{s} \given \mathbf{x}_{\text{obs}})$; and (iii) missing-not-at-random (MNAR), where $p(\mathbf{s} \given \x) = p(\mathbf{s} \given \mathbf{x}_{\text{obs}}, \mathbf{x}_{\text{mis}})$.

The missingness mechanism can be ignored for both MCAR and MAR when learning $p(\mathbf{x}_{\text{obs}}, \mathbf{s})$, but not for MNAR where it depends on $\mathbf{x}_{\text{mis}}$ \citep{ipsen2020not}. We aim to handle all the three cases when performing SBI.

\section{Method} \label{sec:method}

We begin by analyzing the issue of missing data in SBI settings in  \Cref{sec:problemSBI}. We then present RISE --- our method for handling missing data in SBI. \Cref{sec:robustSBI} outlines our learning objective, and \Cref{sec:neural_process} describes how we parameterize the imputation model in \ourmethod using neural processes.

\subsection{Missing data problem in SBI}
\label{sec:problemSBI}

We assume that the simulator can faithfully replicate the true data-generating process (i.e., the simulator is well-specified), however, the data collection mechanism induces missing values in each data point $\x$. As a result, $\x$ contains both observed and missing values,\footnote{Note that during training,  $\x_{\text{obs}}$ and $\x_{\text{mis}}$ are partitions of the simulated data $\x$, while during inference we only observe $\tilde \x_{\text{obs}}$ from the real world.} represented as $\mathbf{x} = (\mathbf{x}_{\text{obs}}, \mathbf{x}_{\text{mis}})$. For instance, $\x = \begin{pmatrix} 0.1 \ 1.2 \ - \ 0.9 \end{pmatrix}$ exemplifies a scenario where a specific coordinate ${x}_i$ is missing (indicated by `$-$'). Naturally, SBI methods cannot operate on missing values, and so imputing $\mathbf{x}_{\text{mis}}$ is necessary before proceeding to inference. However, if the missing values are not imputed accurately, then the corresponding SBI posterior becomes biased (e.g., as observed in \Cref{fig:miss_data} due to constant imputation). We now describe this problem mathematically. 

\begin{definition}[{{SBI posterior under true imputation}}]\label{definition1}
    Let $p_{\text{true}}(\x_{\text{mis}} \given \x_{\text{obs}})$ be the true predictive distribution of the missing values given the observed data. Then, the SBI posterior can be written as
    \begin{align} \label{eq:definition1}
        p_{\text{SBI}}(\theta \given \x_{\text{obs}}) = \int \underbrace{p_{\text{SBI}}(\theta \given \x_{\text{obs}},\x_{\text{mis}})}_\text{Inference} \underbrace{p_{\text{true}}(\x_{\text{mis}} \given \x_{\text{obs}})}_\text{Imputation} d \x_{\text{mis}}.
    \end{align}
\end{definition}
We thus have a distribution over the missing values given $\x_{\text{obs}}$, and the problem of SBI under missing data is formulated as an expectation of the SBI posterior $p_{\text{SBI}}(\theta \given \x_{\text{obs}}, \x_{\text{mis}})$ with respect to $p_{\text{true}}(\x_{\text{mis}} \given \x_{\text{obs}})$, analogous to traditional (likelihood-based) Bayesian inference methods \citep{schafer2000inference,zhou2010note}. 
Therefore, estimating the above expectation requires access to $p_{\text{true}}(\x_{\text{mis}} \given \x_{\text{obs}})$ \citep{raghunathan2001multivariate,gelman1995bayesian}, which is infeasible in most practical cases.
%
\begin{definition}[{{SBI posterior under estimated imputation}}]\label{definition2}
    Let $\hat p(\x_{\text{mis}} \given \x_{\text{obs}})$ denote an estimate of the true imputation model $p_{\text{true}}(\x_{\text{mis}} \given \x_{\text{obs}})$. Then, the corresponding SBI posterior can be written as
    \begin{align} \label{eq:definition2}
        \hat p_{\text{SBI}}(\theta \given \x_{\text{obs}}) = \int p_{\text{SBI}}(\theta \given \x_{\text{obs}},\x_{\text{mis}}) \hat p(\x_{\text{mis}} \given \x_{\text{obs}}) d \x_{\text{mis}}.
    \end{align}
\end{definition}

\begin{proposition}\label{prop:bias}
    If $\hat p(\x_{\text{mis}} \given \x_{\text{obs}})$ is misaligned with $p_{\text{true}}(\x_{\text{mis}} \given \x_{\text{obs}})$, then the estimated SBI posterior $\hat p_{\text{SBI}}(\theta \given \x_{\text{obs}})$ will be biased (in general), i.e., 
    $\mathbb{E}_{\theta \sim p_{\text{SBI}}(\cdot \given \x_{\text{obs}})}[\theta] \neq  \mathbb{E}_{\theta \sim \hat p_{\text{SBI}}(\cdot \given \x_{\text{obs}})} [\theta]$~. 
\end{proposition}
The proof, which follows straightforwardly using \Cref{definition1} and \Cref{definition2}, is given in Appendix \ref{subsec:proof_bias} for completeness. \Cref{prop:bias} says that the bias in the SBI posterior directly comes from the discrepancy between the true imputation model $p_{\text{true}}(\x_{\text{mis}} \given \x_{\text{obs}})$ and the estimated one $\hat p(\x_{\text{mis}} \given \x_{\text{obs}})$. This applies irrespective of the inference method used, and therefore, rather unsurprisingly, the key to reducing this bias is to learn the imputation model as accurately as possible. The rest of this section presents our method, named RISE, which combines the imputation task with SBI to reduce this bias. 

\subsection{Robust SBI under missing data}
\label{sec:robustSBI}

Let $p_{\text{true}}(\theta \given \x_{\text{obs}}, \x_{\text{mis}} )$ be the true posterior given both the observed data and the missing values, i.e., given  $\x = (\x_{\text{obs}}, \x_{\text{mis}})$. Our objective is to estimate the true posterior given only $\x_{\text{obs}}$. That is, we seek to approximate 
$$p_{\text{true}}(\theta \given \x_{\text{obs}}) \triangleq \int p_{\text{true}}(\theta \given \x_{\text{obs}},  \x_{\text{mis}}) p_{\text{true}}(\x_{\text{mis}} \mid \x_{\text{obs}}) d \x_{\text{mis}} ~ = \int p_{\text{true}}(\theta, \x_{\text{mis}} \given \x_{\text{obs}}) d \x_{\text{mis}}~ .$$  
We therefore introduce a family of distributions $r_{\psi} (\theta,\mathbf{x}_{\text{mis}}\mid \mathbf{x}_{\text{obs}})$ parameterized by $\psi$, and propose to solve the following optimization problem 
\begin{equation} \label{eq:original}
          \argmin_{\psi} \,\mathbb{E}_{\mathbf{x}_{\text{obs}} \sim p_{\text{true}}} \, \text{KL}\left[p_{\text{true}}(\theta,\mathbf{x}_{\text{mis}} \mid \mathbf{x}_{\text{obs}} ) \mid \mid \underbrace{r_{\psi}(\theta,\mathbf{x}_{\text{mis}} \mid \mathbf{x}_{\text{obs}})}_{\textbf{(joint imputation and inference)}}\right]~.
\end{equation}
Solving this problem requires access to $p_{\text{true}}(\mathbf{x}_{\text{mis}} \mid \mathbf{x}_{\text{obs}}$), which in most real-world scenarios, we do not have. Since samples for $\mathbf{x}_{\text{mis}}$ are required during training, we need to resort to methods such as variational approximation or expectation maximization.  Here, we adopt a variational approach,  treating $\mathbf{x}_{\text{mis}}$ as latent variables in a probabilistic imputation setting.  Specifically, the imputation network needs to estimate these latents for the inference network to map them to the output space.  Both networks are tightly coupled since the distribution induced by the imputation network shapes the input of the inference network. 

Mathematically, assuming access to only data samples $(\mathbf{x}_{\text{obs}}, \theta) \sim p_{\text{true}}$, we proceed to solving 
\begin{equation} \label{eq:alternative}
          \argmin_{\psi} \,\mathbb{E}_{\mathbf{x}_{\text{obs}} \sim p_{\text{true}}} \, \text{KL}[p_{\text{true}}(\theta \mid \mathbf{x}_{\text{obs}} ) \mid \mid r_{\psi}(\theta \mid \mathbf{x}_{\text{obs}})]~.
\end{equation}
Our next proposition computes a variational lower bound for this objective, which we can maximize efficiently using an encoder-decoder architecture resembling variational autoencoders (VAEs). 
\begin{proposition}[Training objective] 
\label{prop:rise_objective}
The objective in \Cref{eq:alternative} admits a variational lower bound, resulting in the following optimization problem.
\begin{align}
\hat \phi, \hat \varphi & =  \argmin_{\phi, \varphi} -  \,  \mathbb{E}_{(\mathbf{x}_{\text{obs}},\theta) \sim p_{\text{true}}} \mathbb{E}_{\mathbf{x}_{\text{mis}} \sim p(\mathbf{x}_{\text{mis}} \mid \mathbf{x}_{\text{obs}})} \left[\log  \underbrace{\hat{p}_{\varphi}(\mathbf{x}_{\text{mis}} \mid \mathbf{x}_{\text{obs}})}_{\textbf{(imputation)}} + \log \underbrace{q_{\phi}(\theta \mid \mathbf{x}_{\text{obs}}, \mathbf{x}_{\text{mis}})}_{\textbf{(inference)}}\right] \label{eq:rise_objective}
\\
&=\argmin_{\phi, \varphi} \ell_{\ourmethod}(\phi, \varphi) \nonumber,
\end{align}
    where $\ell_{\ourmethod}(\phi, \varphi)$ denotes the loss function for RISE.
\end{proposition}
Therefore, \color{black} we can approximate the true imputation model $p_{\text{true}}(\x_{\text{mis}} \given \x_{\text{obs}})$ using a parametric neural network $\hat{p}_\varphi$, parameterized by its vector of weights and biases $\varphi$, and the SBI posterior given the full dataset $p_{\text{SBI}}(\theta \given \x_{\text{mis}}, \x_{\text{obs}})$ using the conditional density $q_\phi$ as in NPE. %

%
%



The proof of \Cref{prop:rise_objective} is outlined in Appendix \ref{subsec:proof_rise_objective}.  Note that $\ell_{\ourmethod}$ is a general loss which reduces to $\ell_{\text{NPE}}$ when there is no missing data, i.e., $\x = \x_{\text{obs}}$. In case a summary network $\eta_\kappa$ is required before passing the data to $q_\phi$, the joint loss function for \ourmethod can be simply defined as
\begin{equation*}
    \ell_{\ourmethod}(\phi, \varphi, \kappa) \triangleq - \mathbb{E}_{(\x_{\text{obs}}, \theta) \sim p_{\text{true}}, {\x}_{\text{mis}} \sim p( \x_{\text{mis}} \given \x_{\text{obs}} )} \left[ \log q_\phi(\theta \given \eta_\kappa(\x_{\text{obs}},{\x}_{\text{mis}})) + \log \hat{p}_\varphi(\x_{\text{mis}} \given \x_{\text{obs}}) \right].
\end{equation*}
The expectation in \Cref{eq:rise_objective} is taken with respect to the joint distribution of the simulator and the prior (as is standard for SBI methods), and the  variational imputation distribution $p( \x_{\text{mis}} \given \x_{\text{obs}} )$. Note that for simulations in our controlled experiments, we do not need to resort to the variational distribution $p( \x_{\text{mis}} \given \x_{\text{obs}} )$, and can instead generate samples from $p_{\text{true}}( \x_{\text{mis}} \given \x_{\text{obs}} )$  directly by first sampling $\x$ using the simulator, and then partitioning it into $\x_{\text{obs}}$ and $\x_{\text{mis}}$ based on the missingness assumption (i.e. creating the mask $\s$ under MCAR or MAR or MNAR assumption) such that $\varepsilon \%$ portion of the data is missing. The $\x_{\text{mis}}$ values are then used as true labels when comparing against the output of the imputation model $\hat{p}_\varphi$ during training.
This allows us to amortize over instances of real data. In \Cref{sec:neural_process}, we discuss how RISE can be used to amortize over the proportion of missing values $\varepsilon$ in the data.

Using a latent variable representation \citep{kingma2022autoencodingvariationalbayes} for the imputation model, we factorize $\hat{p}_\varphi(\x_{\text{mis}} \given \x_{\text{obs}})$, similarly to  \citet{mattei2019miwae}, as
\begin{align*}
    \hat{p}_\varphi(\x_{\text{mis}} \given \x_{\text{obs}}) =  \int \hat p_{\alpha}(\x_{\text{mis}} \given \tilde{\z})  ~\hat p_{\beta}(\tilde{\z} \given \x_{\text{obs}}) d\tilde{\z}~,
\end{align*}
where $\varphi = (\alpha, \beta)$ are parameters of the imputation model, and $\tilde{\z} = (\z,\s)$ represents both the latent variable $\z$ and the masking variable $\mathbf{s}$, which we can utilize to simulate various missingness environments.  
The conditional distribution of the latent $ \hat p_{\beta}(\tilde{\z} \given \x_{\text{obs}})$ may depend on both the observed and the missing data depending on the different missingness assumptions \citep{little2019statistical}:
\begin{itemize}
    \item MCAR: $\hat p_{\beta}(\tilde{\z} \given \x_{\text{obs}}) = p_{\beta_1}(\z \given \x_{\text{obs}})p_{\beta_2}(\s)$
    \item MAR: $\hat p_{\beta}(\tilde{\z} \given \x_{\text{obs}}) = p_{\beta_1}(\z \given \x_{\text{obs}})p_{\beta_2}(\s \given \x_{\text{obs}})$
    \item MNAR:  $\hat p_{\beta}(\tilde{\z} \given \x_{\text{obs}}) = p_{\beta_1}(\z \given \x_{\text{obs}}) \int  p_{\beta_2}(\s \given \x_{\text{mis}}, \x_{\text{obs}}) p(\x_{\text{mis}} \given \x_{\text{obs}})d \x_{\text{mis}}$~.
\end{itemize}

Note that for the MCAR and MAR cases, we only need the latent $\z$ in order to impute $\x_{\text{mis}}$ \citep{mattei2019miwae}, in which case $\tilde{\z} = \z$. However, in the MNAR case, $\tilde{\z} = (\z, \s)$ as we will explicitly need to account for the missingness mechanism \citep{ipsen2020not}. Hereafter, we continue to denote the latent variable with $\tilde{\z}$ for a general formulation encompassing all the three cases. The pseudocode for training RISE is outlined in \Cref{algo:algorithm}.

\subsection{Learning the imputation model using Neural Process}
\label{sec:neural_process}

We utilize neural processes (NPs, \citet{garnelo2018conditional}) for parameterizing the imputation model $\hat p_\varphi(\x_{\text{mis}} \given \x_{\text{obs}})$. NPs  represent a family of neural network-based meta-learning models
\begin{wrapfigure}[14]{r}{0.24\textwidth}
    \vspace{0pt}
    \hspace{0pt}
    \resizebox{0.2\textwidth}{!}{ \tikz{
     \node[obs] (x) {$\mathbf{x}_{\text{obs}}$};%
     \node[obs,xshift=1cm] (c) {$\mathbf{c}_{\text{obs}}$};%
     
      \plate {plate1} {(x) (c) } {$C$};
    \node[latent,below=of plate1,yshift=-0.03cm] (z) {$\tilde{\mathbf{z}}$}; %
    \node[const, left=of z,xshift=-0.6cm] (beta) {$\beta$}; %
    \node[obs,below=of z,xshift=0.6cm,yshift=-0.06cm] (c_mis) {$\mathbf{c}_{\text{mis}}$};%
     \node[latent,below=of z,xshift=-0.6cm,yshift=-0.06cm] (x_mis) {$\mathbf{x}_{\text{mis}}$}; %
     \node[const,left=of x_mis,yshift=-0.03cm] (alpha) {$\alpha$}; %
   
    \edge {beta} {z}
    \edge {alpha} {x_mis}
    \edge {z} {x_mis}
    \edge {c_mis} {x_mis}
    \edge {plate1} {z}
    \path[->,dashed] (x_mis) edge[bend right=40] node[fill=white, anchor=center, pos=0.7, scale=0.66]{MNAR}(z);
     \plate[inner sep=.30cm] {plate2} {(x_mis)(x) (z) (c)} {$n$};
    }}
     \caption{Plate diagram}
     \label{fig:plate_np}
\end{wrapfigure}
that combine  the flexibility of deep learning with well-calibrated uncertainty estimates and a tractable training objective. These models learn a distribution over predictors given their target positions or locations. We refer the interested reader to Appendix \ref{subsec:npf} for a detailed background. We employ neural processes to model the predictive density over missing values at their specific locations.  

Let $\mathbf c_{\text{mis}} = (c_{\text{mis},1}, \dots, c_{\text{mis},k})$ and $\mathbf c_{\text{obs}} = (c_{\text{obs},1}, \dots, c_{\text{obs},d-k})$ denote the locations pertaining to $\x_{\text{mis}}$ and $\x_{\text{obs}}$, respectively, where $k$ denotes the number of missing values (or the dimensionality of $\x_{\text{mis}}$). Furthermore, let $C = \{ \x_{\text{obs}}, \mathbf{c}_{\text{obs}}\}$ be the observed context set. Then, following latent neural processes \citep{foong2020meta}, we obtain 
\begin{equation} \label{eq:np_lik}
\begin{split}
\hat{p}_\varphi(\x_{\text{mis}} \given \mathbf{c}_{\text{mis}}, C) &= \int \hat p_{\alpha}\big(\x_{\text{mis}} \given \mathbf{c}_{\text{mis}},\tilde{\z}\big) \hat p_{\beta}(\tilde{\z} \given C) d\tilde{\z}  \\ &=  \int \hat p_{\beta}(\tilde{\z} \given C) \prod_{i=1}^{k} \hat p_{\alpha}\big(x_{\text{mis},i} \given c_{\text{mis},i},\tilde{\z}\big)  d\tilde{\z}~. 
\end{split}
\end{equation}

Here we have assumed conditional independence of each $x_{\text{mis},i}$ given $c_{\text{mis},i}$ and $\tilde{\z}$, which allows for the joint distribution to factorize into a product of its marginals. 
Note that this factorization directly inherits the consistency properties from neural processes, as established by \citet{garnelo2018conditional} and \citet{dubois2020npf}, ensuring a consistent distribution representation. The associated plate diagram is given in \Cref{fig:plate_np}. To fully specify the model, we utilize the following:
\begin{itemize}
    \item \textbf{Encoder} $\hat p_{\beta}(\tilde{\z} \given C)$,  which provides a distribution over the latent variables $\tilde{\z}$ having observed the context set $C$. The encoder is parameterized to be permutation invariant to correctly treat $C$ as a set (as required by NPs).
    \item \textbf{Decoder} $ \hat p_{\alpha}(x_{\text{mis},i} \given c_{\text{mis},i},\tilde{\z})$, which provides a predictive distribution over each missing value $x_{\text{mis},i}$ conditioned on $\tilde{\mathbf{z}}$ and the missing location $c_{\text{mis},i}$. In practice, this distribution is assumed to be a Gaussian, and the parameters $\alpha$ denote the predicted mean and variance. 
    
\end{itemize}


The likelihood given in \Cref{eq:np_lik} is not analytically tractable. 
Therefore, following \citet{foong2020meta}, we estimate $\hat{p}_\varphi(\x_{\text{mis}} \given \mathbf{c}_{\text{mis}}, C)$ using $m$ Monte Carlo samples $\tilde{\z}_1, \dots, \tilde{\z}_m \sim \hat p_{\beta}(\tilde{\z} \given C)$ as
\begin{align}
    \log  \hat{p}_\varphi(\x_{\text{mis}} \given \mathbf{c}_{\text{mis}}, C) 
    \approx \log \left( \frac{1}{m} \sum_{j=1}^{m}  \prod_{i=1}^{k} \hat p_{\alpha}\big(x_{\text{mis},i} \given c_{\text{mis},i},\tilde{\z}_j\big)  \right).
\end{align}
This can directly be used with standard optimizers \citep{kingma2014adam} to learn the model parameters. 
\begin{wrapfigure}[13]{r}{0.5\textwidth}
\begin{minipage}{0.45\textwidth}
    \vspace{-22pt}
    \begin{algorithm}[H]
    \caption{RISE (training)}  \label{algo:algorithm}
    \begin{algorithmic}[1]
    \Require  Simulator $p(\cdot \given\theta)$, prior $p(\theta)$, iterations $n_{\mathrm{iter}}$, missingness degree $\varepsilon$
    \State Initialize parameters $\phi,\varphi$ of RISE
    \For{$k = 1, \ldots, n_{\mathrm{iter}}$}
            \State Sample $(\x,\theta) \sim p(\cdot \given \theta) p(\theta)$
            \State Create mask $\s$ wrt $\varepsilon$ and MCAR/MAR/MNAR
            \State Compute $\ell_{\ourmethod}$ using \Cref{eq:rise_objective}
            \State $\phi,\varphi \xleftarrow{} \texttt{optimize}( \ell_{\ourmethod}; \phi,\varphi)$
    \EndFor
    \end{algorithmic}
\end{algorithm}
\end{minipage}
\vspace{-5ex}
\end{wrapfigure}
%

%

As NPs are meta-learning models, we can utilize them to amortize over the proportion of missing values $\varepsilon$. Doing so is beneficial in cases where inference is required on multiple datasets with varying proportions of missing values, so as to avoid re-training for each $\varepsilon$. Assuming $p(\varepsilon)$ to be the distribution of the missingness proportion, we can consider each sample from $p(\varepsilon)$ to be one task when training RISE. Specifically, this can be done by first initializing the parameters of RISE, and then repeating the following: (i) Sample $\varepsilon \sim p(\varepsilon)$, and (ii) Perform Steps 2-7 from \Cref{algo:algorithm}.  We name this variant of our method as RISE-Meta. For each sample from the imputation model, we obtain a posterior distribution via the inference network, thus resulting in an ensemble of posterior distributions across all samples. In \Cref{sec:RISE-Meta}, we test the ability of RISE-Meta to generalize to unknown levels of missing values. 

\section{Related work}

\paragraph{Missing data in SBI.} 
\citet{wang2024missing} attempt to handle missing data by augmenting the missing values with, e.g. zeros or sample mean, and subsequently training NPE with a binary mask indicator, but this approach can lead to biased posterior estimates, as we saw in \Cref{fig:miss_data} and \Cref{sec:problemSBI}.  \citet{wang2022monte,wang2023sbi++} propose imputing missing values by sampling from a kernel density estimate of the training data or using a nearest-neighbor search, and training the NPE model using augmented simulations. However, these approaches neglect the missingness mechanisms, which can distort the relationships between variables \citep{graham2007review} and are not scalable to higher dimensions. \citet{lueckmann2017flexible} learn an imputation model agnostic of the missingness mechanism.  More recently, \citet{gloeckler2024all} have proposed a transformer-based architecture for SBI that can potentially handle conditioning on data with missing values. This method can perform arbitrary conditioning and evaluation, i.e. for a given $\x = [\x_{\text{obs}},\x_{\text{mis}}]$, it first estimates the imputation distribution, i.e. $p(\x_{\text{mis}} \given \x_{\text{obs}})$, and then estimates the posterior distribution $p(\theta \given \x_{\text{obs}},\x_{\text{mis}})$. However, it does not model the mechanism underlying the missing data and is thus not equipped to handle the MNAR settings. In contrast, \ourmethod incorporates the missingness mechanism during its training and is therefore able to estimate the full posterior distribution, accounting for all variables. 


\paragraph{Deep imputation methods.} 
There is a growing body of work on imputing missing data using deep generative models. These include using GANs for missing data under MCAR assumption \citep{yoon2018gain, li2019misgan}, and VAEs under MAR assumption \citep{mattei2019miwae, nazabal2020handling}. Deep generative models have also been studied under MNAR assumption \citep{ghalebikesabi2021deep, gong2021variational, ipsen2020not, ma2021identifiable}. We contribute to this line of work by using latent NPs to handle missing data under all the three missingness assumptions. Instead of learning an imputation model, \citet{smieja2018processing} propose replacing a typical neuron’s response in the first hidden layer by its expected value to process missing data in neural networks.

\section{Experiments}
\label{sec:exp}

In this section, we assess the significance of \ourmethod via detailed empirical investigations. Our first objective is to demonstrate that \ourmethod yields posteriors that are robust to missing values in the data compared to baseline methods (see \Cref{subsec:infer_syn} and \Cref{subsec:infer_real}). Secondly, we aim to test the generalization capability of RISE-Meta in cases where the proportion of missing values in the data is not known \textit{a priori} (\Cref{sec:RISE-Meta}). Thirdly, as learning the imputation model accurately is central to RISE's performance, we aim to validate that employing a NP-based imputation model in RISE yields state-of-the-art results when imputing real-world datasets. Finally, we intend to provide some experimental evidence where learning the inference and imputation components jointly, as is done in RISE, performs better than learning them separately. 

Our experiments are organized as follows. We first provide results on SBI benchmarks in Sections \ref{subsec:infer_syn}, \ref{subsec:infer_real}, and \ref {sec:RISE-Meta}. In \Cref{sec:ablation}, we report our ablation studies to evaluate the imputation performance of RISE on real-world bioactivity datasets. 

\paragraph{Performance metrics.} We evaluate the accuracy of the posterior using the following metrics: (i) the nominal log posterior probability of true parameters (NLPP), (ii) the classifier two-sample test (C2ST) score \citep{lopez-paz2017revisiting}, and (iii) the maximum
mean discrepancy (MMD) \citep{gretton2012kernel}. The MMD and C2ST metrics are computed between the posterior samples obtained under missing data (either using \ourmethod or the baseline methods) and samples from a reference NPE posterior under no missing data. We use a radial basis function kernel for computing the MMD, and set its lengthscale using the median heuristic \citep{gretton2012kernel} on the reference posterior samples.  

\paragraph{Baselines.} We evaluate RISE's performance against baselines derived from NPE \citep{greenberg2019automatic}. These include the mask-based method proposed by \citet{wang2024missing}, and NPE-NN that combines NPE with a feed-forward neural network for joint training and imputation \citep{lueckmann2017flexible}. While NPE-NN shares RISE's joint training paradigm, it performs single imputation rather than the multiple imputation approach used in RISE. We also compare against Simformer \citep{gloeckler2024all}, a recent diffusion and transformer-based approach for posterior estimation. 
\begin{table}[!t]
\centering
\caption{NLPP and C2ST metrics under MCAR and MNAR scenarios, with missing value proportion $\varepsilon$. \ourmethod demonstrates superior posterior estimation performance. For MNAR scenarios, the proportion of missing values averages below $\varepsilon$ due to self-censoring (details in \Cref{sec:creating_mask}). Note that Simformer results are unavailable for Ricker and OUP due to the lack of official implementation.}
\resizebox{\textwidth}{!}{
\begin{tabular}{@{}llcccccccccc@{}}
\toprule
& \multirow{2}{*}{Dataset} & \multirow{2}{*}{$\epsilon$} & \multicolumn{4}{c}{NLPP}                                     &        &   \multicolumn{4}{c}{C2ST}                \\ \cmidrule{4-7} \cmidrule{9-12} 
 && & NPE-NN           & Wang et al. & Simformer        & RISE             &  & NPE-NN & Wang et al. & Simformer & RISE \\ \midrule 
\multirow{12}{*}{\rotatebox{90}{MCAR}} &\multirow{3}{*}{GLU}    & $10 \%$ & $-2.51 \pm 0.11$ &  $-2.50 \pm 0.10$            & $-2.45 \pm 0.12$ & $\mathbf{-2.31} \pm 0.10$ &  & $0.87\pm 0.01$   &   $0.87\pm 0.01$          &    $0.85\pm 0.01$       & $\mathbf{0.83}\pm 0.01$ \\ 
 &                       & $25 \%$ & $-3.92 \pm 0.11$ &   $\mathbf{-3.54} \pm 0.17$          & $-3.65 \pm 0.17$ & $-3.71 \pm 0.11$ &  & $0.90\pm 0.01$   &    $0.92\pm 0.01$         &    $0.91\pm 0.01$       & $\mathbf{0.89}\pm 0.01$ \\ 
 &                       & $60 \%$ & $-6.37 \pm 0.12$ & $-6.52 \pm 0.17$            & $-6.62 \pm 0.27$ & $\mathbf{-6.21} \pm 0.11$ &  & $0.98\pm 0.01$   &     $0.97\pm 0.01$        &   $0.96\pm 0.01$        & $\mathbf{0.93}\pm 0.01$ \\ 
\cmidrule{2-12}
&\multirow{3}{*}{GLM}    & $10 \%$ & $-6.57 \pm 0.13$ &     $-7.10 \pm 0.11$        & $-6.47 \pm 0.16$ & $\mathbf{-6.32} \pm 0.15$ &  & $0.84\pm 0.01$   &       $0.86\pm 0.01$      &    $0.84\pm 0.01$       & $\mathbf{0.80}\pm 0.01$ \\ 
&                        & $25 \%$ & $-7.72 \pm 0.16$ &        $-7.84 \pm 0.17$     & $-7.37 \pm 0.13$ & $\mathbf{-7.22} \pm 0.17$ &  & $0.93\pm 0.01$   &    $0.94\pm 0.01$          &   $0.92\pm 0.01$        & $\mathbf{0.91}\pm 0.01$ \\  
&                        & $60 \%$& $-9.02 \pm 0.17$ &   $-8.97 \pm 0.15$          & $-8.93 \pm 0.18$ & $\mathbf{-8.71} \pm 0.14$ &  & $0.99\pm 0.01$   &     $0.99\pm 0.01$        &   $0.98\pm 0.01$        & $\mathbf{0.97}\pm 0.01$ \\  \cmidrule{2-12}
&\multirow{3}{*}{Ricker} & $10 \%$ & $-4.90 \pm 0.16$ &   $-4.74 \pm 0.31$          & -                & $\mathbf{-4.20} \pm 0.09$ &  & $0.94\pm 0.01$   &     $0.93\pm 0.01$        & -          & $\mathbf{0.90}\pm 0.01$ \\ 
&                        & $25 \%$ & $-4.94 \pm 0.17$ &  $-5.14 \pm 0.27$           &    -              & $\mathbf{-4.64} \pm 0.15$ &  & $0.96\pm 0.01$   &    $0.95\pm 0.01$         & -          & $\mathbf{0.92}\pm 0.01$ \\ 
&                        & $60 \%$ & $-4.97 \pm 0.17$ &     $-5.24 \pm 0.11$        & -                 & $\mathbf{-4.72} \pm 0.17$ &  & $0.97\pm 0.01$   &    $0.99\pm 0.01$         & -          & $\mathbf{0.94}\pm 0.01$ \\ \cmidrule{2-12}
&\multirow{3}{*}{OUP}    & $10 \%$ & $-2.25 \pm 0.18$ &     $-2.37 \pm 0.18$        & -                 & $\mathbf{-2.09} \pm 0.11$ &  & $0.89\pm 0.01$   &    $0.88\pm 0.01$         & -          & $\mathbf{0.87}\pm 0.01$ \\ 
 &                       & $25 \%$ & $-2.74 \pm 0.18$ &     $-2.55 \pm 0.13$        & -                 & $\mathbf{-2.43} \pm 0.15$ &  & $0.90\pm 0.01$   &    $0.91 \pm 0.01$         & -          & $\mathbf{0.89}\pm 0.01$ \\  
 &                       & $60 \%$ & $-2.87 \pm 0.19$ &    $-2.75 \pm 0.17$         & -                 & $\mathbf{-2.52} \pm 0.11$ &  & $0.95\pm 0.01$   &    $0.94 \pm 0.01$         & -          & $\mathbf{0.93}\pm 0.01$ \\ \midrule
 \multirow{12}{*}{\rotatebox{90}{MNAR}} &\multirow{3}{*}{GLU}    & $10 \%$ & $ -2.35 \pm 0.10 $ &         $ -2.42 \pm 0.17$     & $ -2.15\pm0.10 $ & $-\mathbf{1.90} \pm 0.09$ &  &  $0.89 \pm 0.01$  &      $0.88 \pm 0.01$       &   $0.87 \pm 0.01$       & $\mathbf{0.85} \pm 0.01$  \\ 
 &                       & $25 \%$ & $-3.31 \pm 0.17 $ &   $ -3.67 \pm 0.12$          & $ -\mathbf{3.12} \pm 0.12$ & $ -3.26 \pm 0.10$ &  & $0.92\pm 0.01$   &   $0.93\pm 0.01$           &    $0.91\pm 0.01$     &  $\mathbf{0.88}\pm 0.01$ \\ 
 &                       & $60 \%$ & $ -5.97 \pm 0.19$ &   $ -6.03\pm 0.11$          & $ -6.02\pm 0.12$ & $ -\mathbf{5.80}\pm 0.27$ &  & $0.96\pm 0.01$   &   $0.95\pm 0.01$          &    $0.93\pm 0.01$       & $\mathbf{0.92}\pm 0.01$ \\ 
\cmidrule{2-12}
&\multirow{3}{*}{GLM}    & $10 \%$ & $ -6.05 \pm 0.27 $ &      $ -5.98 \pm 0.22 $        & $-6.17 \pm 0.18 $ & $ -\mathbf{5.82} \pm 0.11 $ &  &  $0.89\pm 0.01$ &      $0.90\pm 0.01$       &  $0.87\pm 0.01$         & $\mathbf{0.85}\pm 0.01$ \\ 
&                        & $25 \%$ & $-6.47 \pm 0.14 $ &    $ -6.51 \pm 0.32 $         & $ -6.57 \pm 0.14$ & $-\mathbf{6.12} \pm 0.15 $ &  &  $0.94\pm 0.01$  &    $0.95\pm 0.01$         &  $0.92\pm 0.01$         & $\mathbf{0.89}\pm 0.01$ \\  
&                        & $60 \%$& $-7.78 \pm 0.37 $ &    $ -8.38 \pm 0.12 $         & $ -7.56 \pm 0.15 $ & $ -\mathbf{7.11} \pm 0.17 $ &  & $0.97\pm 0.01$   &    $0.98\pm 0.01$          &    $0.96\pm 0.01$       & $\mathbf{0.95}\pm 0.01$ \\  \cmidrule{2-12}
&\multirow{3}{*}{Ricker} & $10 \%$ & $-4.67 \pm 0.24$ &   $ -4.35 \pm 0.13$          & -                 & $ -\mathbf{4.10} \pm 0.18 $ &  &  $0.94\pm 0.01$  &       $0.94\pm 0.01$      & -          & $\mathbf{0.92}\pm 0.01$ \\ 
&                        & $25 \%$ & $ -4.91 \pm 0.20 $ &    $ -5.05 \pm 0.18$         & -                 & $ -\mathbf{4.75} \pm 0.23$ &  &   $0.95\pm 0.01$ &     $0.96\pm 0.01$        & -          &  $\mathbf{0.93}\pm 0.01$\\ 
&                        & $60 \%$ & $ -5.25 \pm 0.21$ & $ -5.12 \pm 0.16$             & -                 & $ -\mathbf{4.82} \pm 0.26$ &  &  $0.97\pm 0.01$ &    $0.99\pm 0.01$         & -          & $\mathbf{0.95}\pm 0.01$ \\ \cmidrule{2-12}
&\multirow{3}{*}{OUP}    & $10 \%$ & $-2.21 \pm 0.13 $ &      $ -2.32 \pm 0.18$        & -                 & $ -\mathbf{2.10} \pm 0.12$ &  &  $0.93\pm 0.01$  &       $0.92\pm 0.01$      & -          & $\mathbf{0.88} \pm 0.01 $ \\ 
 &                       & $25 \%$ & $-2.42 \pm 0.17 $ &    $ -2.57 \pm 0.11$          & -                 & $ -\mathbf{2.24} \pm 0.17$  &  &  $0.97 \pm 0.01$  &     $0.95 \pm 0.01$        & -          &  $\mathbf{0.93} \pm 0.01 $\\  
 &                       & $60 \%$ & $ -2.92 \pm 0.15 $ &     $ -2.79 \pm 0.21$         & -                 & $ -\mathbf{2.47} \pm 0.21$  &  & $0.99\pm 0.01$ &        $0.99\pm 0.01$     & -          & $\mathbf{0.97} \pm 0.01$  \\ \bottomrule
\end{tabular}}

\label{tab:nlpp_c2st}
\end{table}




\paragraph{Implementation.} \ourmethod is implemented in PyTorch~\citep{paszke2019pytorch} and utilizes the same training configuration as the competing baselines (see \Cref{sec:network_param} for details). We take ${\varepsilon} \in \{10\%,25\%,60\%\}$ to test performance from low to high missingness scenarios. We adopt the  masking approach as described in \citet{mattei2019miwae} and \citet{ipsen2020not} for MCAR and MNAR respectively. Specifically, for MCAR we randomly mask $\varepsilon\%$ of the data, and for MNAR we use $\varepsilon$ to compute a  masking probability, which is then used to mask data according to their values. This \textit{self-censoring} approach is described in \Cref{sec:creating_mask}, and leads to a missingness proportion less than (or equal to) $\varepsilon$. We set a simulation budget of $n=1000$ for all the SBI experiments, and take 1000 samples from the posterior distributions to compute the MMD, C2ST and NLPP. The performance is evaluated over 10 random runs. For further details, see \Cref{subsec:implement}.

\subsection{Performance on SBI benchmarks}\label{subsec:infer_syn}
We evaluate the performance of RISE in settings with missing data using four common benchmark models from the SBI literature, namely, (i) Ricker model: a two parameter simulator from population genetics \citep{Wood2010}; (ii) Ornstein-Uhlenbeck process (OUP): a two parameter stochastic differential equation model \citep{Chen2021neural}; (iii) Generalized Linear Model (GLM): a 10 parameter model with Bernoulli observations; and (iv) Gaussian Linear Uniform (GLU): a 10-dimensional Gaussian model with the mean vector as the parameter and a fixed covariance. The models are described in \Cref{sec:model_descriptions}, and the prior distributions we used are reported in \Cref{subsubsec:prior}.

The results for NLPP, C2ST are shown in \Cref{tab:nlpp_c2st} and MMD in \Cref{tab:mmd}, comparing performance across varying missingness levels $\varepsilon$ under both MCAR and MNAR conditions. We observe that \ourmethod achieves the lowest values of C2ST across missingness types, thus outperforming the baselines in estimating the posterior distributions. As $\varepsilon$ increases, the gap between RISE and the baselines increases, indicating that RISE is able to better handle high missingness levels in the data. As a sanity check, we also investigate the imputation capability of \ourmethod. \Cref{fig:imp_mse} shows that RISE achieves better imputation, which then naturally translates to robust posterior estimation. The difference in performance is more stark in the MNAR case, as expected, since the baseline methods do not explicitly model the missingness mechanism. 

\begin{figure}[!t]
    \centering
    \includegraphics[width=\textwidth]{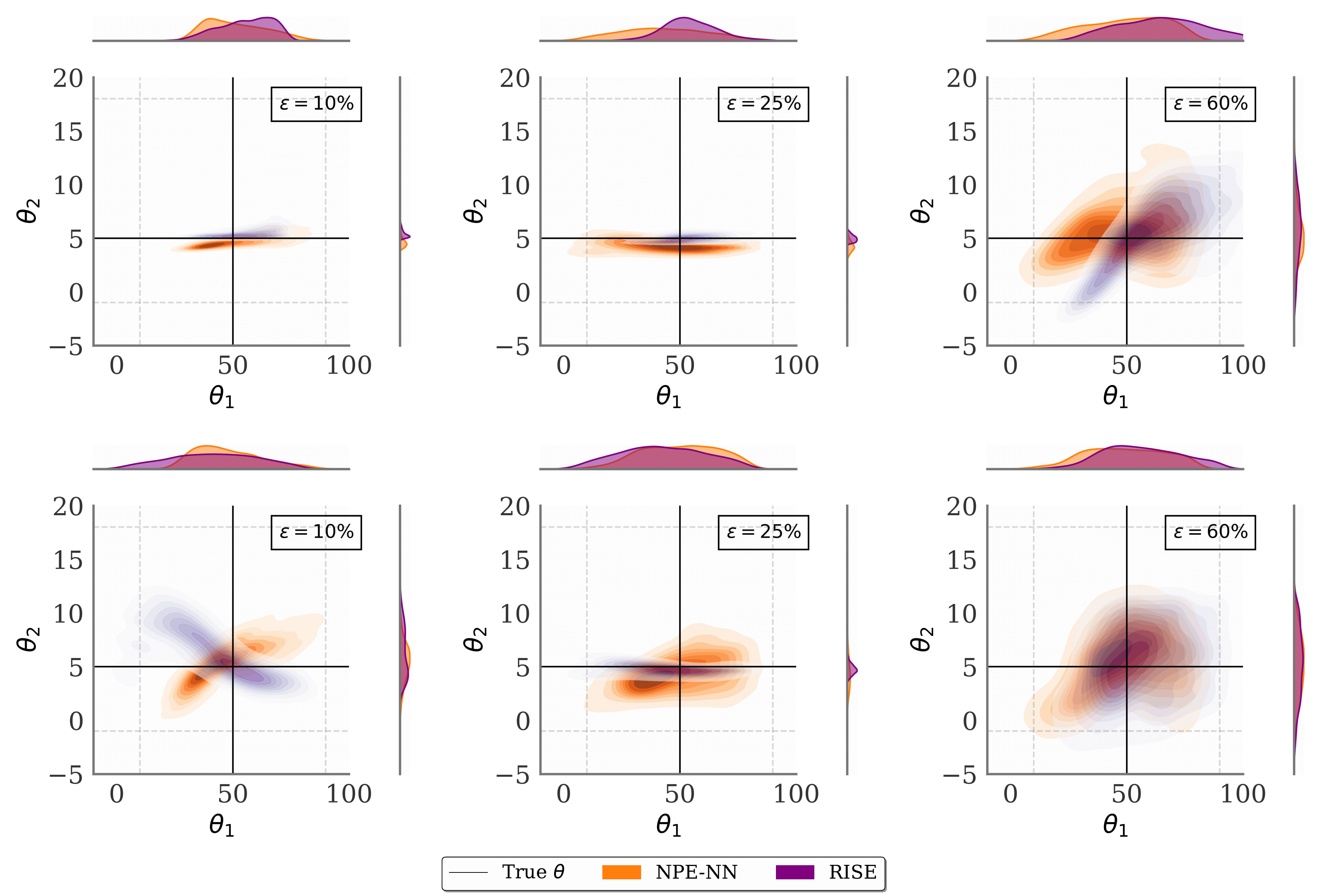}
    \caption{{Posterior estimates for the Hodgkin-Huxley model under MCAR (\textit{top row}) and MNAR (\textit{bottom row}) with varying proportions of missing values in the data (denoted by $\varepsilon$).} The posteriors obtained from \ourmethod stay close to the true parameter (denoted by the black lines) for all values of $\varepsilon$, while those from the baseline methods move further away as $\varepsilon$ increases.}
    \label{fig:hh}
\end{figure}

\subsection{Hodgkin-Huxley model}\label{subsec:infer_real}

We now apply RISE on a real-world computational neuroscience simulator \citep{hodgkin1952quantitative}, namely the Hodgkin-Huxley model, which is a popular example in the SBI literature \citep{Lueckmann2017, gao2024generalized, gloeckler2023adversarial}.  
The aim is to infer the posterior over two parameters given the data of dimension 1200 (see \Cref{sec:model_descriptions} for the model description). 

We set uniform priors and perform inference under different values of $\varepsilon$ and missingness assumptions, similar to \Cref{subsec:infer_syn}. \Cref{fig:hh} shows that RISE's posteriors are robust to increasing proportions of missing values as they stay around the true parameter value as compared to NPE-NN. We also evaluate the expected coverage of the posterior in \Cref{subsec:coverage}, which demonstrates that \ourmethod produces conservative posterior approximations and achieves better calibration than NPE-NN. 


\subsection{Generalizing across unknown levels of missingness}
\label{sec:RISE-Meta}
\begin{wrapfigure}[16]{r}{0.43\textwidth}
    \vspace{0pt}
    \includegraphics[width=0.43\textwidth]{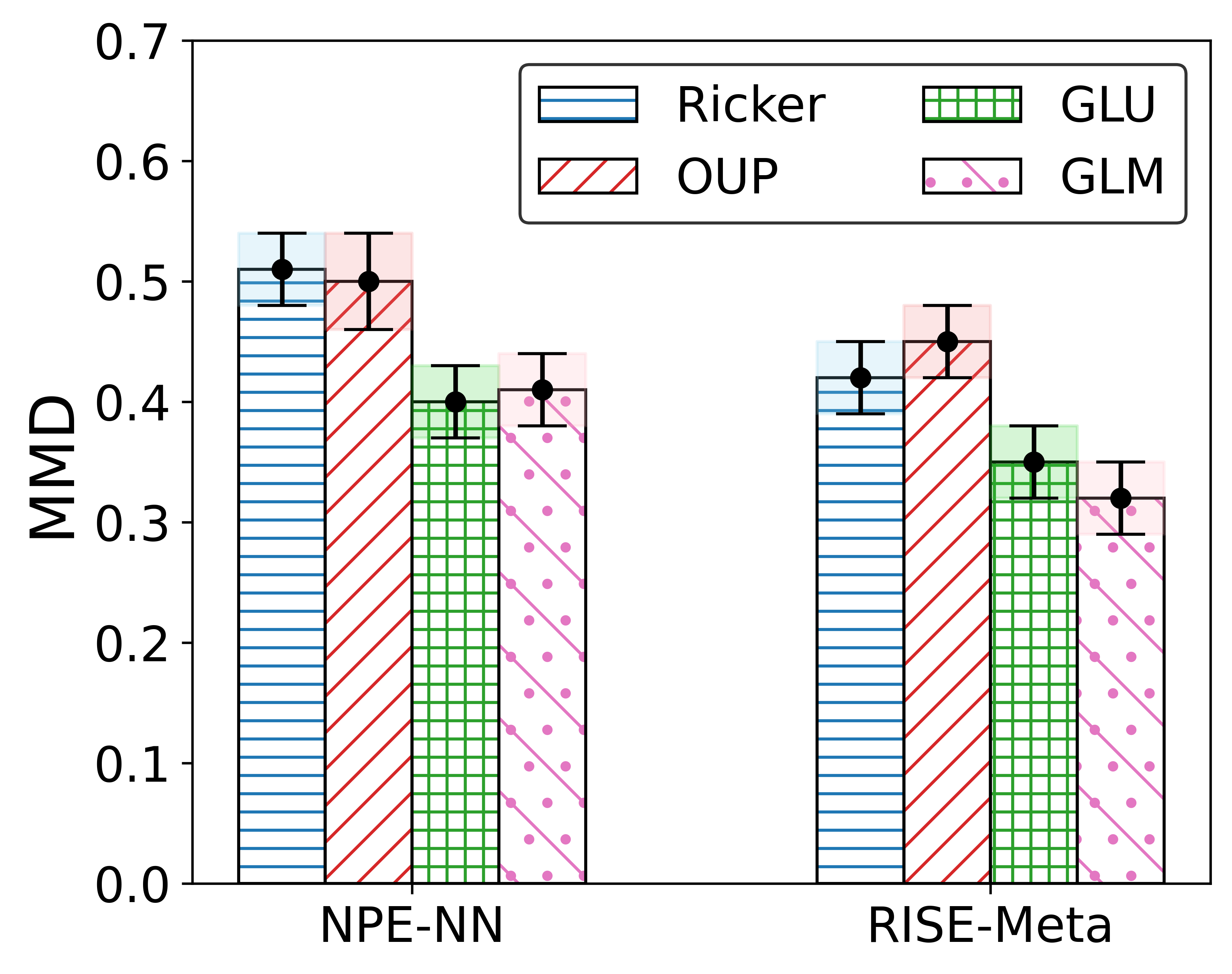}
     \caption{Generalizing over missingness.}
     \label{fig:meta}
\end{wrapfigure}

Next, we test the generalization capability of our method to unknown levels of missing values. We perform meta-learning over different proportions of missing values ${\varepsilon}$ in the dataset (termed RISE-Meta). For training RISE-Meta, we take the distribution of $p({\varepsilon})$ to be an equiprobable discrete distribution on the set $\{10\%, 25\%, 60\%\}$. We also train NPE-NN with a missingness degree of $60\%$ as a baseline. We evaluate all the methods over 100 samples of varying missingness proportion $\varepsilon \sim \mathcal{U}([0,1])$. \Cref{fig:meta} shows the MMD results on GLM, GLU, Ricker and OUP tasks.  We observe that RISE-Meta achieves the lowest MMD values for both the tasks, thus demonstrating its ability to better generalize to unknown levels of missing values in the data. 

\subsection{Ablation studies}
\label{sec:ablation}
\paragraph{Imputation performance on real-world datasets.} 
We now look at how the neural process-based imputation model in \ourmethod performs on real-world datasets. The 
task is to predict and impute bioactivity data on Adrenergic receptor assays \citep{whitehead2019imputation} and Kinase assays \citep{martin2017profile} from the field of drug discovery.  The Kinase test data consists of outliers, unlike the training data, which makes imputation challenging. We can therefore use such data to assess the generalization capabilities of RISE.
We compare the RISE imputation method to other methods from this field such as QSAR \citep{cherkasov2014qsar}, Conduit\footnote{Since, the official implementation is unavailable, we use the re-implementation provided here: \url{https://github.com/PenelopeJones/neural_processes}.} \citep{whitehead2019imputation}, and Collective Matrix Factorization (CMF) \citep{singh2008relational}. We also include a standard deep neural network (DNN) and a vanilla neural process as baselines. Table \ref{tab:abl} (left) reports the coefficient of determination $R^2$ \citep{wright1921correlation} between the true and the predicted assays. We observe that \ourmethod achieves state-of-the-art results in these tasks, demonstrating the efficacy of the neural processes-based imputation model. 

\begin{table}[!hbt]
     \caption{\textbf{Ablation studies.} (\textit{left}) $R^2$ scores ($\uparrow$) on the bioactivity datasets. (\textit{right}) RMSE ($\downarrow$) across different proportions of missingness ($\varepsilon$) for comparing the effect of joint versus separate learning. }
    \label{tab:abl}
    \vskip 0.05in
    \hspace{5pt}
    \resizebox{0.40\textwidth}{!}{
    \begin{tabular}{lcc}
        \toprule
        Method & Adrenergic  & Kinase   \\
        \midrule
        QSAR & \textcolor{gray}{(N/A)} & -0.19 $\pm$ \textcolor{gray}{0.01}\\
        CMF & 0.59 $\pm$ \textcolor{gray}{0.02} & -0.11 $\pm$ \textcolor{gray}{0.01} \\
        DNN & 0.60 $\pm$ \textcolor{gray}{0.05}  & 0.11 $\pm$ \textcolor{gray}{0.01} \\
        NP & 0.61 $\pm$ \textcolor{gray}{0.03}  & 0.17 $\pm$ \textcolor{gray}{0.04} \\
        Conduilt & 0.62 $\pm$ \textcolor{gray}{0.04}  & 0.22 $\pm$ \textcolor{gray}{0.03} \\
        CNP & 0.65 $\pm$ \textcolor{gray}{0.04}  & 0.24 $\pm$ \textcolor{gray}{0.02} \\
        \midrule
        RISE & \textbf{0.67} $\pm$ \textcolor{gray}{0.03}  & \textbf{0.26} $\pm$ \textcolor{gray}{0.03} \\
        \bottomrule
    \end{tabular}}
    \hspace{10pt}
    \resizebox{0.47\textwidth}{!}{
    \begin{tabular}{llcc}
        \toprule
        \textbf{Missigness ($\epsilon$)} & \textbf{Method} & \textbf{GLM} & \textbf{GLU}   \\
        \midrule
         \multirow{3}{*}{$10\%$}&NPE-RF-Sep&  0.69 $\pm$\textcolor{gray}{0.03}&   0.44 $\pm$\textcolor{gray}{0.02}\\
         &RISE-Sep& 0.67 $\pm$\textcolor{gray}{0.03}   & 0.43 $\pm$\textcolor{gray}{0.02}\\
         &RISE&\textbf{0.65} $\pm$\textcolor{gray}{0.04}   & \textbf{0.41}$\pm$\textcolor{gray}{0.01}\\
        \midrule
         \multirow{3}{*}{$25\%$}&NPE-RF-Sep&  1.02 $\pm$\textcolor{gray}{0.05}&  0.48 $\pm$\textcolor{gray}{0.02}\\
         &RISE-Sep& 0.99 $\pm$\textcolor{gray}{0.03}    &  0.45 $\pm$\textcolor{gray}{0.02}\\
         &RISE& \textbf{0.93} $\pm$\textcolor{gray}{0.06}  &\textbf{0.43} $\pm$\textcolor{gray}{0.02}\\
        \midrule
         \multirow{3}{*}{$60\%$}&NPE-RF-Sep&  1.34 $\pm$\textcolor{gray}{0.10}& 0.64   $\pm$\textcolor{gray}{0.02}\\
         &RISE-Sep& 1.31 $\pm$\textcolor{gray}{0.03}    & 0.58 $\pm$\textcolor{gray}{0.03}\\
         &RISE& \textbf{1.27} $\pm$\textcolor{gray}{0.01} & \textbf{0.56} $\pm$\textcolor{gray}{0.03}\\
        
                \bottomrule
    \end{tabular}}
    \label{tab:imput_meta}
    
\end{table}

\paragraph{Joint vs separate learning.} This experiment involves investigating the impact of training the imputation and the inference model in \ourmethod jointly (as we proposed) versus separately (termed RISE-Sep). We also include another baseline termed NPE-RF-Sep where a random forest (RF) model is first used for imputation, followed by NPE. Table \ref{tab:imput_meta} (right) reports the RMSE values on GLM and GLU tasks for different missingness proportion $\varepsilon$. We observe that training the imputation and inference networks jointly yields improvement in performance over training them separately.

We report the results from additional ablation studies for runtime comparisons, flow architecture, and simulation budget in \Cref{sec:add_abl}.
\section{Conclusion and Limitations}

We analyzed the problem of performing SBI under missing data and showed that inaccurately imputing the missing values may lead to bias in the resulting posterior distributions. We then proposed RISE as a method that aims to reduce this bias under different notions of the underlying missingness mechanism. RISE combines the inference network of NPE with an imputation model based on neural processes (NPs) to achieve robustness to missing data whilst being amortized. Additionally, RISE can be trained in a meta-learning manner over the proportion of missing values in the data, thus allowing for amortization across datasets with varying levels of missingness. While \ourmethod offers substantial advantages, there are limitations to address. RISE inherits the issues of NPE and may yield posteriors that are not well-calibrated (see, e.g., \citet{hermans2022a}). Moreover, the normality assumption in NPs may exhibit limited expressivity in practice when learning complex imputation distributions.

\section*{Acknowledgements}
 YV and VG acknowledge support from the Research Council of Finland for the “Human-steered next-generation machine learning for reviving drug design” project (grant decision 342077). VG also acknowledges the support from Jane and Aatos Erkko Foundation (grant 7001703) for “Biodesign: Use of artificial intelligence in enzyme design for synthetic biology”. AB is supported by the Research Council of Finland grant no. 362534. The experiments were performed using resources provided by the Aalto University Science-IT project and CSC – IT Center for Science,
Finland. YV thanks Priscilla Ong for highlighting relevant related works on deep imputation models and for insightful discussions on various missingness mechanisms.

\bibliography{sample}

\begin{thebibliography}{87}
\providecommand{\natexlab}[1]{#1}
\providecommand{\url}[1]{\texttt{#1}}
\expandafter\ifx\csname urlstyle\endcsname\relax
  \providecommand{\doi}[1]{doi: #1}\else
  \providecommand{\doi}{doi: \begingroup \urlstyle{rm}\Url}\fi

\bibitem[Bharti et~al.(2022{\natexlab{a}})Bharti, Briol, and Pedersen]{Bharti2022}
Ayush Bharti, Francois-Xavier Briol, and Troels Pedersen.
\newblock A general method for calibrating stochastic radio channel models with kernels.
\newblock \emph{{IEEE} Transactions on Antennas and Propagation}, 70\penalty0 (6):\penalty0 3986--4001, 2022{\natexlab{a}}.
\newblock \doi{10.1109/tap.2021.3083761}.

\bibitem[Bharti et~al.(2022{\natexlab{b}})Bharti, Filstroff, and Kaski]{Bharti22ICML}
Ayush Bharti, Louis Filstroff, and Samuel Kaski.
\newblock Approximate {B}ayesian computation with domain expert in the loop.
\newblock In \emph{International Conference on Machine Learning}, volume 162, pages 1893--1905, 2022{\natexlab{b}}.

\bibitem[Briol et~al.(2019)Briol, Barp, Duncan, and Girolami]{Briol2019MMD}
F-X. Briol, A.~Barp, A.~B. Duncan, and M.~Girolami.
\newblock {Statistical inference for generative models with maximum mean discrepancy}.
\newblock \emph{arXiv:1906.05944}, 2019.

\bibitem[Casale et~al.(2018)Casale, Dalca, Saglietti, Listgarten, and Fusi]{casale2018gaussian}
Francesco~Paolo Casale, Adrian Dalca, Luca Saglietti, Jennifer Listgarten, and Nicolo Fusi.
\newblock Gaussian process prior variational autoencoders.
\newblock \emph{Advances in neural information processing systems}, 31, 2018.

\bibitem[Chan et~al.(2018)Chan, Perrone, Spence, Jenkins, Mathieson, and Song]{chan2018likelihood}
Jeffrey Chan, Valerio Perrone, Jeffrey Spence, Paul Jenkins, Sara Mathieson, and Yun Song.
\newblock A likelihood-free inference framework for population genetic data using exchangeable neural networks.
\newblock \emph{Advances in neural information processing systems}, 31, 2018.

\bibitem[Chen et~al.(2021)Chen, Zhang, Gutmann, Courville, and Zhu]{Chen2021neural}
Yanzhi Chen, Dinghuai Zhang, Michael~U. Gutmann, Aaron Courville, and Zhanxing Zhu.
\newblock Neural approximate sufficient statistics for implicit models.
\newblock In \emph{International Conference on Learning Representations}, 2021.

\bibitem[Cherkasov et~al.(2014)Cherkasov, Muratov, Fourches, Varnek, Baskin, Cronin, Dearden, Gramatica, Martin, Todeschini, et~al.]{cherkasov2014qsar}
Artem Cherkasov, Eugene~N Muratov, Denis Fourches, Alexandre Varnek, Igor~I Baskin, Mark Cronin, John Dearden, Paola Gramatica, Yvonne~C Martin, Roberto Todeschini, et~al.
\newblock Qsar modeling: where have you been? where are you going to?
\newblock \emph{Journal of medicinal chemistry}, 57\penalty0 (12):\penalty0 4977--5010, 2014.

\bibitem[Collier et~al.(2020)Collier, Nazabal, and Williams]{collier2020vaes}
Mark Collier, Alfredo Nazabal, and Christopher~KI Williams.
\newblock Vaes in the presence of missing data.
\newblock \emph{arXiv preprint arXiv:2006.05301}, 2020.

\bibitem[Cranmer et~al.(2020)Cranmer, Brehmer, and Louppe]{cranmer2020frontier}
Kyle Cranmer, Johann Brehmer, and Gilles Louppe.
\newblock The frontier of simulation-based inference.
\newblock \emph{Proceedings of the National Academy of Sciences}, 117\penalty0 (48):\penalty0 30055--30062, 2020.

\bibitem[Dax et~al.(2021)Dax, Green, Gair, Macke, Buonanno, and Schölkopf]{Dax2021}
Maximilian Dax, Stephen~R. Green, Jonathan Gair, Jakob~H. Macke, Alessandra Buonanno, and Bernhard Schölkopf.
\newblock Real-time gravitational wave science with neural posterior estimation.
\newblock \emph{Physical Review Letters}, 127\penalty0 (24):\penalty0 241103, December 2021.
\newblock ISSN 1079-7114.
\newblock \doi{10.1103/physrevlett.127.241103}.

\bibitem[Dellaporta et~al.(2022)Dellaporta, Knoblauch, Damoulas, and Briol]{Dellaporta2022}
Charita Dellaporta, Jeremias Knoblauch, Theodoros Damoulas, and Francois-Xavier Briol.
\newblock Robust bayesian inference for simulator-based models via the mmd posterior bootstrap.
\newblock In \emph{International Conference on Artificial Intelligence and Statistics}, volume 151, pages 943--970, 2022.

\bibitem[Dubois et~al.(2020)Dubois, Gordon, and Foong]{dubois2020npf}
Yann Dubois, Jonathan Gordon, and Andrew~YK Foong.
\newblock Neural process family.
\newblock \url{http://yanndubs.github.io/Neural-Process-Family/}, September 2020.

\bibitem[Durkan et~al.(2019)Durkan, Bekasov, Murray, and Papamakarios]{durkan2019neural}
Conor Durkan, Artur Bekasov, Iain Murray, and George Papamakarios.
\newblock Neural spline flows.
\newblock \emph{Advances in neural information processing systems}, 32, 2019.

\bibitem[Foong et~al.(2020)Foong, Bruinsma, Gordon, Dubois, Requeima, and Turner]{foong2020meta}
Andrew Foong, Wessel Bruinsma, Jonathan Gordon, Yann Dubois, James Requeima, and Richard Turner.
\newblock Meta-learning stationary stochastic process prediction with convolutional neural processes.
\newblock \emph{Advances in Neural Information Processing Systems}, 33:\penalty0 8284--8295, 2020.

\bibitem[Fortuin et~al.(2020)Fortuin, Baranchuk, Ratsch, and Mandt]{fortuin2020gp}
Vincent Fortuin, Dmitry Baranchuk, Gunnar Ratsch, and Stephan Mandt.
\newblock Gp-vae: Deep probabilistic time series imputation.
\newblock In \emph{International conference on artificial intelligence and statistics}, pages 1651--1661. PMLR, 2020.

\bibitem[Frazier et~al.(2020)Frazier, Robert, and Rousseau]{Frazier2020ABC}
David~T. Frazier, Christian~P. Robert, and Judith Rousseau.
\newblock Model misspecification in approximate bayesian computation: consequences and diagnostics.
\newblock \emph{Journal of the Royal Statistical Society: Series B (Statistical Methodology)}, 82\penalty0 (2):\penalty0 421--444, 2020.
\newblock \doi{10.1111/rssb.12356}.

\bibitem[Fujisawa et~al.(2021)Fujisawa, Teshima, Sato, and Sugiyama]{2021Fujisawa}
Masahiro Fujisawa, Takeshi Teshima, Issei Sato, and Masashi Sugiyama.
\newblock $\gamma$-abc: Outlier-robust approximate bayesian computation based on a robust divergence estimator.
\newblock In \emph{International Conference on Artificial Intelligence and Statistics}, volume 130, pages 1783--1791, 2021.

\bibitem[Gao et~al.(2023)Gao, Deistler, and Macke]{gao2024generalized}
Richard Gao, Michael Deistler, and Jakob~H Macke.
\newblock Generalized bayesian inference for scientific simulators via amortized cost estimation.
\newblock \emph{Advances in Neural Information Processing Systems}, 36, 2023.

\bibitem[Garnelo et~al.(2018)Garnelo, Rosenbaum, Maddison, Ramalho, Saxton, Shanahan, Teh, Rezende, and Eslami]{garnelo2018conditional}
Marta Garnelo, Dan Rosenbaum, Christopher Maddison, Tiago Ramalho, David Saxton, Murray Shanahan, Yee~Whye Teh, Danilo Rezende, and SM~Ali Eslami.
\newblock Conditional neural processes.
\newblock In \emph{International conference on machine learning}, pages 1704--1713. PMLR, 2018.

\bibitem[Geffner et~al.(2023)Geffner, Papamakarios, and Mnih]{geffner2023compositional}
Tomas Geffner, George Papamakarios, and Andriy Mnih.
\newblock Compositional score modeling for simulation-based inference.
\newblock In \emph{International Conference on Machine Learning}, pages 11098--11116. PMLR, 2023.

\bibitem[Gelman et~al.(1995)Gelman, Carlin, Stern, and Rubin]{gelman1995bayesian}
Andrew Gelman, John~B Carlin, Hal~S Stern, and Donald~B Rubin.
\newblock \emph{Bayesian data analysis}.
\newblock Chapman and Hall/CRC, 1995.

\bibitem[Ghalebikesabi et~al.(2021{\natexlab{a}})Ghalebikesabi, Cornish, Holmes, and Kelly]{ghalebikesabi2021deep}
Sahra Ghalebikesabi, Rob Cornish, Chris Holmes, and Luke Kelly.
\newblock Deep generative missingness pattern-set mixture models.
\newblock In \emph{International conference on artificial intelligence and statistics}, pages 3727--3735. PMLR, 2021{\natexlab{a}}.

\bibitem[Ghalebikesabi et~al.(2021{\natexlab{b}})Ghalebikesabi, Cornish, Kelly, and Holmes]{ghalebikesabi2021deepgenerativepatternsetmixture}
Sahra Ghalebikesabi, Rob Cornish, Luke~J. Kelly, and Chris Holmes.
\newblock Deep generative pattern-set mixture models for nonignorable missingness, 2021{\natexlab{b}}.

\bibitem[Gloeckler et~al.(2023)Gloeckler, Deistler, and Macke]{gloeckler2023adversarial}
Manuel Gloeckler, Michael Deistler, and Jakob~H Macke.
\newblock Adversarial robustness of amortized bayesian inference.
\newblock \emph{arXiv preprint arXiv:2305.14984}, 2023.

\bibitem[Gloeckler et~al.(2024)Gloeckler, Deistler, Weilbach, Wood, and Macke]{gloeckler2024all}
Manuel Gloeckler, Michael Deistler, Christian~Dietrich Weilbach, Frank Wood, and Jakob~H. Macke.
\newblock All-in-one simulation-based inference.
\newblock In Ruslan Salakhutdinov, Zico Kolter, Katherine Heller, Adrian Weller, Nuria Oliver, Jonathan Scarlett, and Felix Berkenkamp, editors, \emph{Proceedings of the 41st International Conference on Machine Learning}, volume 235 of \emph{Proceedings of Machine Learning Research}, pages 15735--15766. PMLR, 21--27 Jul 2024.

\bibitem[Gong et~al.(2021)Gong, Hajimirsadeghi, He, Durand, and Mori]{gong2021variational}
Yu~Gong, Hossein Hajimirsadeghi, Jiawei He, Thibaut Durand, and Greg Mori.
\newblock Variational selective autoencoder: Learning from partially-observed heterogeneous data.
\newblock In \emph{International Conference on Artificial Intelligence and Statistics}, pages 2377--2385. PMLR, 2021.

\bibitem[Graham et~al.(2007)Graham, Olchowski, and Gilreath]{graham2007review}
JW~Graham, AE~Olchowski, and TD~Gilreath.
\newblock Review: A gentle introduction to imputation of missing values.
\newblock \emph{Prev. Sci}, 8:\penalty0 206--213, 2007.

\bibitem[Greenberg et~al.(2019)Greenberg, Nonnenmacher, and Macke]{greenberg2019automatic}
David Greenberg, Marcel Nonnenmacher, and Jakob Macke.
\newblock Automatic posterior transformation for likelihood-free inference.
\newblock In \emph{International Conference on Machine Learning}, pages 2404--2414. PMLR, 2019.

\bibitem[Gretton et~al.(2012)Gretton, Borgwardt, Rasch, Sch{\"o}lkopf, and Smola]{gretton2012kernel}
Arthur Gretton, Karsten~M Borgwardt, Malte~J Rasch, Bernhard Sch{\"o}lkopf, and Alexander Smola.
\newblock A kernel two-sample test.
\newblock \emph{The Journal of Machine Learning Research}, 13\penalty0 (1):\penalty0 723--773, 2012.

\bibitem[Hermans et~al.(2022)Hermans, Delaunoy, Rozet, Wehenkel, Begy, and Louppe]{hermans2022a}
Joeri Hermans, Arnaud Delaunoy, Fran{\c{c}}ois Rozet, Antoine Wehenkel, Volodimir Begy, and Gilles Louppe.
\newblock A crisis in simulation-based inference? beware, your posterior approximations can be unfaithful.
\newblock \emph{Transactions on Machine Learning Research}, 2022.
\newblock ISSN 2835-8856.
\newblock URL \url{https://openreview.net/forum?id=LHAbHkt6Aq}.

\bibitem[Hodgkin and Huxley(1952)]{hodgkin1952quantitative}
Alan~L Hodgkin and Andrew~F Huxley.
\newblock A quantitative description of membrane current and its application to conduction and excitation in nerve.
\newblock \emph{The Journal of physiology}, 117\penalty0 (4):\penalty0 500, 1952.

\bibitem[Huang et~al.(2023)Huang, Bharti, Souza, Acerbi, and Kaski]{huang2024learning}
Daolang Huang, Ayush Bharti, Amauri Souza, Luigi Acerbi, and Samuel Kaski.
\newblock Learning robust statistics for simulation-based inference under model misspecification.
\newblock \emph{Advances in Neural Information Processing Systems}, 36, 2023.

\bibitem[Ipsen et~al.(2020)Ipsen, Mattei, and Frellsen]{ipsen2020not}
Niels~Bruun Ipsen, Pierre-Alexandre Mattei, and Jes Frellsen.
\newblock not-miwae: Deep generative modelling with missing not at random data.
\newblock \emph{arXiv preprint arXiv:2006.12871}, 2020.

\bibitem[Kasak et~al.(2024)Kasak, Deja, Karwowska, Jakubowska, Graczykowski, and Janik]{kasak2024machine}
Mi{l}osz Kasak, Kamil Deja, Maja Karwowska, Monika Jakubowska, {L}ukasz Graczykowski, and Ma{l}gorzata Janik.
\newblock Machine-learning-based particle identification with missing data.
\newblock \emph{The European Physical Journal C}, 84\penalty0 (7):\penalty0 691, 2024.

\bibitem[Kelly et~al.(2024)Kelly, Nott, Frazier, Warne, and Drovandi]{kelly2024misspecificationrobust}
Ryan~P. Kelly, David~J Nott, David~Tyler Frazier, David~J Warne, and Christopher Drovandi.
\newblock Misspecification-robust sequential neural likelihood for simulation-based inference.
\newblock \emph{Transactions on Machine Learning Research}, 2024.
\newblock ISSN 2835-8856.

\bibitem[Kingma(2013)]{kingma2022autoencodingvariationalbayes}
Diederik~P Kingma.
\newblock Auto-encoding variational bayes.
\newblock \emph{arXiv preprint arXiv:1312.6114}, 2013.

\bibitem[Kingma(2014)]{kingma2014adam}
Diederik~P Kingma.
\newblock Adam: A method for stochastic optimization.
\newblock \emph{arXiv preprint arXiv:1412.6980}, 2014.

\bibitem[Kypraios et~al.(2017)Kypraios, Neal, and Prangle]{Kypraios2017}
Theodore Kypraios, Peter Neal, and Dennis Prangle.
\newblock A tutorial introduction to bayesian inference for stochastic epidemic models using approximate bayesian computation.
\newblock \emph{Mathematical Biosciences}, 287:\penalty0 42--53, May 2017.
\newblock ISSN 0025-5564.
\newblock \doi{10.1016/j.mbs.2016.07.001}.

\bibitem[Li et~al.(2019)Li, Jiang, and Marlin]{li2019misgan}
Steven Cheng-Xian Li, Bo~Jiang, and Benjamin Marlin.
\newblock Misgan: Learning from incomplete data with generative adversarial networks.
\newblock \emph{International Conference on Learning Representations}, 2019.

\bibitem[Linhart et~al.(2024)Linhart, Cardoso, Gramfort, Corff, and Rodrigues]{linhart2024diffusion}
Julia Linhart, Gabriel~Victorino Cardoso, Alexandre Gramfort, Sylvain~Le Corff, and Pedro~LC Rodrigues.
\newblock Diffusion posterior sampling for simulation-based inference in tall data settings.
\newblock \emph{arXiv preprint arXiv:2404.07593}, 2024.

\bibitem[Little and Rubin(2019)]{little2019statistical}
Roderick~JA Little and Donald~B Rubin.
\newblock \emph{Statistical analysis with missing data}, volume 793.
\newblock John Wiley \& Sons, 2019.

\bibitem[Lopez-Paz and Oquab(2017)]{lopez-paz2017revisiting}
David Lopez-Paz and Maxime Oquab.
\newblock Revisiting classifier two-sample tests.
\newblock In \emph{International Conference on Learning Representations}, 2017.
\newblock URL \url{https://openreview.net/forum?id=SJkXfE5xx}.

\bibitem[Lueckmann et~al.(2017{\natexlab{a}})Lueckmann, Goncalves, Bassetto, {\"O}cal, Nonnenmacher, and Macke]{lueckmann2017flexible}
Jan-Matthis Lueckmann, Pedro~J Goncalves, Giacomo Bassetto, Kaan {\"O}cal, Marcel Nonnenmacher, and Jakob~H Macke.
\newblock Flexible statistical inference for mechanistic models of neural dynamics.
\newblock \emph{Advances in neural information processing systems}, 30, 2017{\natexlab{a}}.

\bibitem[Lueckmann et~al.(2017{\natexlab{b}})Lueckmann, Gon\c{c}alves, Bassetto, \"{O}cal, Nonnenmacher, and Macke]{Lueckmann2017}
Jan-Matthis Lueckmann, Pedro~J. Gon\c{c}alves, Giacomo Bassetto, Kaan \"{O}cal, Marcel Nonnenmacher, and Jakob~H. Macke.
\newblock Flexible statistical inference for mechanistic models of neural dynamics.
\newblock In \emph{Advances in Neural Information Processing Systems (NIPS)}, page 1289–1299, 2017{\natexlab{b}}.

\bibitem[Lueckmann et~al.(2021)Lueckmann, Boelts, Greenberg, Goncalves, and Macke]{lueckmann2021benchmarking}
Jan-Matthis Lueckmann, Jan Boelts, David Greenberg, Pedro Goncalves, and Jakob Macke.
\newblock Benchmarking simulation-based inference.
\newblock In \emph{International conference on artificial intelligence and statistics}, pages 343--351. PMLR, 2021.

\bibitem[Luken et~al.(2021)Luken, Padhy, and Wang]{luken2021missing}
Kieran~J Luken, Rabina Padhy, and X~Rosalind Wang.
\newblock Missing data imputation for galaxy redshift estimation.
\newblock \emph{arXiv preprint arXiv:2111.13806}, 2021.

\bibitem[Luo et~al.(2018)Luo, Cai, Zhang, Xu, et~al.]{luo2018multivariate}
Yonghong Luo, Xiangrui Cai, Ying Zhang, Jun Xu, et~al.
\newblock Multivariate time series imputation with generative adversarial networks.
\newblock \emph{Advances in neural information processing systems}, 31, 2018.

\bibitem[Ma and Zhang(2021)]{ma2021identifiable}
Chao Ma and Cheng Zhang.
\newblock Identifiable generative models for missing not at random data imputation.
\newblock \emph{Advances in Neural Information Processing Systems}, 34:\penalty0 27645--27658, 2021.

\bibitem[Martin et~al.(2017)Martin, Polyakov, Tian, and Perez]{martin2017profile}
Eric~J Martin, Valery~R Polyakov, Li~Tian, and Rolando~C Perez.
\newblock Profile-qsar 2.0: kinase virtual screening accuracy comparable to four-concentration ic50s for realistically novel compounds.
\newblock \emph{Journal of chemical information and modeling}, 57\penalty0 (8):\penalty0 2077--2088, 2017.

\bibitem[Mattei and Frellsen(2019)]{mattei2019miwae}
Pierre-Alexandre Mattei and Jes Frellsen.
\newblock Miwae: Deep generative modelling and imputation of incomplete data sets.
\newblock In \emph{International Conference on Machine Learning}, pages 4413--4423. PMLR, 2019.

\bibitem[Muzellec et~al.(2020)Muzellec, Josse, Boyer, and Cuturi]{muzellec2020missingdataimputationusing}
Boris Muzellec, Julie Josse, Claire Boyer, and Marco Cuturi.
\newblock Missing data imputation using optimal transport, 2020.

\bibitem[Nazabal et~al.(2020)Nazabal, Olmos, Ghahramani, and Valera]{nazabal2020handling}
Alfredo Nazabal, Pablo~M Olmos, Zoubin Ghahramani, and Isabel Valera.
\newblock Handling incomplete heterogeneous data using vaes.
\newblock \emph{Pattern Recognition}, 107:\penalty0 107501, 2020.

\bibitem[Oksendal(2013)]{oksendal2013stochastic}
Bernt Oksendal.
\newblock \emph{Stochastic differential equations: an introduction with applications}.
\newblock Springer Science \& Business Media, 2013.

\bibitem[Ong et~al.(2024)Ong, Haussmann, and Lahdesmaki]{ong2024learning}
Priscilla Ong, Manuel Haussmann, and Harri Lahdesmaki.
\newblock Learning high-dimensional mixed models via amortized variational inference.
\newblock In \emph{ICML 2024 Workshop on Structured Probabilistic Inference {\&} Generative Modeling}, 2024.

\bibitem[Papamakarios and Murray(2016)]{papamakarios2016fast}
George Papamakarios and Iain Murray.
\newblock Fast $\varepsilon$-free inference of simulation models with bayesian conditional density estimation.
\newblock \emph{Advances in neural information processing systems}, 29, 2016.

\bibitem[Papamakarios et~al.(2017)Papamakarios, Pavlakou, and Murray]{papamakarios2017masked}
George Papamakarios, Theo Pavlakou, and Iain Murray.
\newblock Masked autoregressive flow for density estimation.
\newblock \emph{Advances in neural information processing systems}, 30, 2017.

\bibitem[Papamakarios et~al.(2019)Papamakarios, Sterratt, and Murray]{papamakarios2019sequential}
George Papamakarios, David Sterratt, and Iain Murray.
\newblock Sequential neural likelihood: Fast likelihood-free inference with autoregressive flows.
\newblock In \emph{The 22nd International Conference on Artificial Intelligence and Statistics}, pages 837--848. PMLR, 2019.

\bibitem[Papamakarios et~al.(2021)Papamakarios, Nalisnick, Rezende, Mohamed, and Lakshminarayanan]{papamakarios2021normalizing}
George Papamakarios, Eric Nalisnick, Danilo~Jimenez Rezende, Shakir Mohamed, and Balaji Lakshminarayanan.
\newblock Normalizing flows for probabilistic modeling and inference.
\newblock \emph{Journal of Machine Learning Research}, 22\penalty0 (57):\penalty0 1--64, 2021.

\bibitem[Paszke et~al.(2019)Paszke, Gross, Massa, Lerer, Bradbury, Chanan, Killeen, Lin, Gimelshein, Antiga, et~al.]{paszke2019pytorch}
Adam Paszke, Sam Gross, Francisco Massa, Adam Lerer, James Bradbury, Gregory Chanan, Trevor Killeen, Zeming Lin, Natalia Gimelshein, Luca Antiga, et~al.
\newblock Pytorch: An imperative style, high-performance deep learning library.
\newblock \emph{Advances in neural information processing systems}, 32, 2019.

\bibitem[Pesonen et~al.(2023)Pesonen, Simola, K{\"o}hn-Luque, Vuollekoski, Lai, Frigessi, Kaski, Frazier, Maneesoonthorn, Martin, et~al.]{pesonen2023abc}
Henri Pesonen, Umberto Simola, Alvaro K{\"o}hn-Luque, Henri Vuollekoski, Xiaoran Lai, Arnoldo Frigessi, Samuel Kaski, David~T Frazier, Worapree Maneesoonthorn, Gael~M Martin, et~al.
\newblock Abc of the future.
\newblock \emph{International Statistical Review}, 91\penalty0 (2):\penalty0 243--268, 2023.

\bibitem[Pospischil et~al.(2008)Pospischil, Toledo-Rodriguez, Monier, Piwkowska, Bal, Fr{\'e}gnac, Markram, and Destexhe]{pospischil2008minimal}
Martin Pospischil, Maria Toledo-Rodriguez, Cyril Monier, Zuzanna Piwkowska, Thierry Bal, Yves Fr{\'e}gnac, Henry Markram, and Alain Destexhe.
\newblock Minimal hodgkin--huxley type models for different classes of cortical and thalamic neurons.
\newblock \emph{Biological cybernetics}, 99:\penalty0 427--441, 2008.

\bibitem[Radev et~al.(2020)Radev, Mertens, Voss, Ardizzone, and Köthe]{Radev2022}
Stefan~T. Radev, Ulf~K. Mertens, Andreas Voss, Lynton Ardizzone, and Ullrich Köthe.
\newblock Bayesflow: Learning complex stochastic models with invertible neural networks, 2020.
\newblock URL \url{https://arxiv.org/abs/2003.06281}.

\bibitem[Raghunathan et~al.(2001)Raghunathan, Lepkowski, Van~Hoewyk, Solenberger, et~al.]{raghunathan2001multivariate}
Trivellore~E Raghunathan, James~M Lepkowski, John Van~Hoewyk, Peter Solenberger, et~al.
\newblock A multivariate technique for multiply imputing missing values using a sequence of regression models.
\newblock \emph{Survey methodology}, 27\penalty0 (1):\penalty0 85--96, 2001.

\bibitem[Ramchandran et~al.(2021)Ramchandran, Tikhonov, Kujanp{\"a}{\"a}, Koskinen, and L{\"a}hdesm{\"a}ki]{ramchandran2021longitudinal}
Siddharth Ramchandran, Gleb Tikhonov, Kalle Kujanp{\"a}{\"a}, Miika Koskinen, and Harri L{\"a}hdesm{\"a}ki.
\newblock Longitudinal variational autoencoder.
\newblock In \emph{International Conference on Artificial Intelligence and Statistics}, pages 3898--3906. PMLR, 2021.

\bibitem[Riesselman et~al.(2018)Riesselman, Ingraham, and Marks]{Riesselman2018}
Adam~J. Riesselman, John~B. Ingraham, and Debora~S. Marks.
\newblock Deep generative models of genetic variation capture the effects of mutations.
\newblock \emph{Nature Methods}, 15\penalty0 (10):\penalty0 816--822, September 2018.
\newblock ISSN 1548-7105.
\newblock \doi{10.1038/s41592-018-0138-4}.

\bibitem[Rubin(1976)]{rubin1976inference}
Donald~B Rubin.
\newblock Inference and missing data.
\newblock \emph{Biometrika}, 63\penalty0 (3):\penalty0 581--592, 1976.

\bibitem[Schafer and Schenker(2000)]{schafer2000inference}
Joseph~L Schafer and Nathaniel Schenker.
\newblock Inference with imputed conditional means.
\newblock \emph{Journal of the American Statistical Association}, 95\penalty0 (449):\penalty0 144--154, 2000.

\bibitem[Schmitt et~al.(2023)Schmitt, B{\"u}rkner, K{\"o}the, and Radev]{schmitt2023detecting}
Marvin Schmitt, Paul-Christian B{\"u}rkner, Ullrich K{\"o}the, and Stefan~T Radev.
\newblock Detecting model misspecification in amortized bayesian inference with neural networks.
\newblock In \emph{DAGM German Conference on Pattern Recognition}, pages 541--557. Springer, 2023.

\bibitem[Sinelnikov et~al.(2024)Sinelnikov, Haussmann, and L{\"a}hdesm{\"a}ki]{sinelnikov2024latent}
Maksim Sinelnikov, Manuel Haussmann, and Harri L{\"a}hdesm{\"a}ki.
\newblock Latent variable model for high-dimensional point process with structured missingness.
\newblock \emph{arXiv preprint arXiv:2402.05758}, 2024.

\bibitem[Singh and Gordon(2008)]{singh2008relational}
Ajit~P Singh and Geoffrey~J Gordon.
\newblock Relational learning via collective matrix factorization.
\newblock In \emph{Proceedings of the 14th ACM SIGKDD international conference on Knowledge discovery and data mining}, pages 650--658, 2008.

\bibitem[Sisson(2018)]{Sisson2018}
Scott~A. Sisson.
\newblock \emph{Handbook of Approximate {B}ayesian Computation}.
\newblock Chapman and Hall/{CRC}, 2018.

\bibitem[{S}mieja et~al.(2018){S}mieja, Struski, Tabor, Zieli{n}ski, and Spurek]{smieja2018processing}
Marek {S}mieja, {L}ukasz Struski, Jacek Tabor, Bartosz Zieli{n}ski, and Przemys{l}aw Spurek.
\newblock Processing of missing data by neural networks.
\newblock \emph{Advances in neural information processing systems}, 31, 2018.

\bibitem[Tejero-Cantero et~al.(2020)Tejero-Cantero, Boelts, Deistler, Lueckmann, Durkan, Gonçalves, Greenberg, and Macke]{tejero-cantero2020sbi}
Alvaro Tejero-Cantero, Jan Boelts, Michael Deistler, Jan-Matthis Lueckmann, Conor Durkan, Pedro~J. Gonçalves, David~S. Greenberg, and Jakob~H. Macke.
\newblock sbi: A toolkit for simulation-based inference.
\newblock \emph{Journal of Open Source Software}, 5\penalty0 (52):\penalty0 2505, 2020.
\newblock \doi{10.21105/joss.02505}.

\bibitem[Van~Buuren and Groothuis-Oudshoorn(2011)]{van2011mice}
Stef Van~Buuren and Karin Groothuis-Oudshoorn.
\newblock mice: Multivariate imputation by chained equations in r.
\newblock \emph{Journal of statistical software}, 45:\penalty0 1--67, 2011.

\bibitem[Vo et~al.(2024)Vo, Zhao, Le, Bonilla, and Phung]{vo2024optimal}
Vy~Vo, He~Zhao, Trung Le, Edwin~V Bonilla, and Dinh Phung.
\newblock Optimal transport for structure learning under missing data.
\newblock \emph{arXiv preprint arXiv:2402.15255}, 2024.

\bibitem[Wang et~al.(2022)Wang, Leja, Villar, and Speagle]{wang2022monte}
Bingjie Wang, Joel Leja, Ashley Villar, and Joshua~S Speagle.
\newblock Monte carlo techniques for addressing large errors and missing data in simulation-based inference.
\newblock \emph{arXiv preprint arXiv:2211.03747}, 2022.

\bibitem[Wang et~al.(2023)Wang, Leja, Villar, and Speagle]{wang2023sbi++}
Bingjie Wang, Joel Leja, V~Ashley Villar, and Joshua~S Speagle.
\newblock Sbi++: Flexible, ultra-fast likelihood-free inference customized for astronomical applications.
\newblock \emph{The Astrophysical Journal Letters}, 952\penalty0 (1):\penalty0 L10, 2023.

\bibitem[Wang et~al.(2024)Wang, Hasenauer, and Sch{\"a}lte]{wang2024missing}
Zijian Wang, Jan Hasenauer, and Yannik Sch{\"a}lte.
\newblock Missing data in amortized simulation-based neural posterior estimation.
\newblock \emph{PLOS Computational Biology}, 20\penalty0 (6):\penalty0 e1012184, 2024.

\bibitem[Ward et~al.(2022)Ward, Cannon, Beaumont, Fasiolo, and Schmon]{Ward2022robust}
Daniel Ward, Patrick Cannon, Mark Beaumont, Matteo Fasiolo, and Sebastian~M Schmon.
\newblock Robust neural posterior estimation and statistical model criticism.
\newblock In \emph{Advances in Neural Information Processing Systems}, 2022.

\bibitem[Whitehead et~al.(2019)Whitehead, Irwin, Hunt, Segall, and Conduit]{whitehead2019imputation}
Thomas~M Whitehead, Benedict~WJ Irwin, P~Hunt, Matthew~D Segall, and Gareth~John Conduit.
\newblock Imputation of assay bioactivity data using deep learning.
\newblock \emph{Journal of chemical information and modeling}, 59\penalty0 (3):\penalty0 1197--1204, 2019.

\bibitem[Wood(2010)]{Wood2010}
Simon~N. Wood.
\newblock Statistical inference for noisy nonlinear ecological dynamic systems.
\newblock \emph{Nature}, 466\penalty0 (7310):\penalty0 1102--1104, 2010.
\newblock \doi{10.1038/nature09319}.

\bibitem[Wright(1921)]{wright1921correlation}
Sewall Wright.
\newblock Correlation and causation.
\newblock \emph{Journal of agricultural research}, 20\penalty0 (7):\penalty0 557, 1921.

\bibitem[Yoon et~al.(2018)Yoon, Jordon, and Schaar]{yoon2018gain}
Jinsung Yoon, James Jordon, and Mihaela Schaar.
\newblock Gain: Missing data imputation using generative adversarial nets.
\newblock In \emph{International Conference on Machine Learning}, pages 5689--5698. PMLR, 2018.

\bibitem[Yoon and Sull(2020)]{yoon2020gamin}
Seongwook Yoon and Sanghoon Sull.
\newblock Gamin: Generative adversarial multiple imputation network for highly missing data.
\newblock In \emph{Proceedings of the IEEE/CVF conference on computer vision and pattern recognition}, pages 8456--8464, 2020.

\bibitem[Zaheer et~al.(2017)Zaheer, Kottur, Ravanbakhsh, Poczos, Salakhutdinov, and Smola]{zaheer2017deep}
Manzil Zaheer, Satwik Kottur, Siamak Ravanbakhsh, Barnabas Poczos, Russ~R Salakhutdinov, and Alexander~J Smola.
\newblock Deep sets.
\newblock \emph{Advances in neural information processing systems}, 30, 2017.

\bibitem[Zhao et~al.(2023)Zhao, Sun, Dezfouli, and Bonilla]{zhao2023transformed}
He~Zhao, Ke~Sun, Amir Dezfouli, and Edwin~V Bonilla.
\newblock Transformed distribution matching for missing value imputation.
\newblock In \emph{International Conference on Machine Learning}, pages 42159--42186. PMLR, 2023.

\bibitem[Zhou and Reiter(2010)]{zhou2010note}
Xiang Zhou and Jerome~P Reiter.
\newblock A note on bayesian inference after multiple imputation.
\newblock \emph{The American Statistician}, 64\penalty0 (2):\penalty0 159--163, 2010.

\end{thebibliography}
\bibliographystyle{plainnat}

\newpage

\appendix
\section{Appendix}
\Cref{subsec:dis} discusses the aspect of handling multiple observations and model misspecfication in the context of \ourmethod. In \Cref{subsec:proof}, we present the proofs for \Cref{prop:bias} and \Cref{prop:rise_objective}. \Cref{subsec:npf} provides a background on neural processes, and \Cref{subsec:implement} presents the implementation details for the experiments of \Cref{sec:exp}. \Cref{sec:add_results} contains additional metrics, coverage plots and visulaizations. \Cref{sec:add_abl} reports the results from additional ablation studies.

\subsection{Discussion} \label{subsec:dis}

\paragraph{Handling multiple observations.} Although so far we have focused on the single observation case where we have one data vector $\mathbf{x}$ for each $\theta$, RISE can straightforwardly be extended to the multiple observations case where we obtain $\x^{(1:m)} = (\x_1, \dots, \x_m)$ for each $\theta$. Then, $\x^{(1:m)} = (\x^{(1:m)}_{\text{obs}}, \x^{(1:m)}_{\text{mis}})$, and the objective for RISE becomes 
\small
\begin{equation*}
    \argmin_{\phi, \varphi, \kappa} -  \,  \mathbb{E}_{(\mathbf{x}^{(1:m)}_{\text{obs}},\theta) \sim p_{\text{true}}} \mathbb{E}_{\mathbf{x}^{(1:m)}_{\text{mis}} \sim \prod_{i=1}^m p(\mathbf{x}^{(i)}_{\text{mis}} \mid \mathbf{x}^{(i)}_{\text{obs}})} \left[\dfrac{1}{m} \sum_{i=1}^m\log  \underbrace{\hat{p}_{\varphi}(\mathbf{x}^{(i)}_{\text{mis}} \mid \mathbf{x}^{(i)}_{\text{obs}})}_{\textbf{(imputation)}} + \log \underbrace{q_{\phi}(\theta \mid \eta_\kappa (\mathbf{x}^{1:m}_{\text{obs}}, \mathbf{x}^{1:m}_{\text{mis}}))}_{\textbf{(inference)}}\right].
\end{equation*}
\normalsize
Note that we can summarize the data using the network $\eta_\kappa$ (for instance, a deep set \citep{zaheer2017deep}) before passing the data into the inference network, which is standard practice when using NPE with multiple observations \citep{chan2018likelihood}. Alternatively, one could use recent extensions based on score estimation \citep{geffner2023compositional, linhart2024diffusion} as well.

\paragraph{Handling model misspecification.} We conjecture that replacing the inference network in RISE from the usual NPE to a robust variant such as the method of \citet{Ward2022robust} or \citet{huang2024learning} would help in addressing model misspecification issues. It would be an interesting avenue for future research to see how to train these robust NPE methods jointly with the imputation network of RISE, and how effective such an approach is. One way is to assume a certain error model over the observed data $\mathbf{x}$, corrupt the data to $\tilde{\mathbf{x}}$ by adding a Gaussian noise, and infer the correct $\theta$ via the inference network. This can be formulated as
\small
\begin{align}
    \mathrm{argmin}_{\phi,\varphi} - \mathbb{E}_{ 
 (\mathbf{x}_{\text{obs}}, \theta) \sim p(\mathbf{x}_{\text{obs}}, \theta), \tilde{\mathbf{x}}_{\text{obs}} \sim \mathcal{N}(\mathbf{x}_{\text{obs}},\sigma^2),  {\mathbf{\tilde{x}}}_{\text{mis}} \sim p_{\text{true}}( \mathbf{\tilde{x}}_{\text{mis}} \mid \mathbf{\tilde{x}}_{\text{obs}}, \theta )} \left[ \log \hat{p}_\varphi(\mathbf{\tilde{x}}_{\text{mis}} \mid \mathbf{\tilde{x}}_{\text{obs}}) + \log q_\phi(\theta \mid \mathbf{\tilde{x}}_{\text{obs}},{\mathbf{\tilde{x}}}_{\text{mis}})  \right].
\end{align}
\normalsize
Moreover, our method can also be readily extended to incorporate prior mis-specification:
\small
\begin{align}
    \mathrm{argmin}_{\phi,\varphi} - \mathbb{E}_{ 
 (\mathbf{x}_{\text{obs}}, \theta) \sim p(\mathbf{x}_{\text{obs}}, \theta), \tilde{\theta} \sim \mathcal{N}(\theta,\sigma^2),  {\mathbf{\tilde{x}}}_{\text{mis}} \sim p_{\text{true}}( \mathbf{\tilde{x}}_{\text{mis}} \mid \mathbf{\tilde{x}}_{\text{obs}}, \tilde{\theta} )} \left[ \log \hat{p}_\varphi(\mathbf{\tilde{x}}_{\text{mis}} \mid \mathbf{\tilde{x}}_{\text{obs}}) + \log q_\phi(\theta \mid \mathbf{\tilde{x}}_{\text{obs}},{\mathbf{\tilde{x}}}_{\text{mis}}) \right].
\end{align}
\normalsize
\color{black}
\subsection{Proofs}\label{subsec:proof}

\subsubsection{Proof for \Cref{prop:bias}} 
\label{subsec:proof_bias}

\begin{proof}
    Using \Cref{eq:definition1} and \Cref{eq:definition2}, we note that
    \begin{align*}
        &\mathbb{E}_{\theta \sim p_{\text{SBI}}(\theta \given \x_{\text{obs}})}[\theta] -  \mathbb{E}_{\theta \sim \hat p_{\text{SBI}}(\theta \given \x_{\text{obs}})} [\theta]  \\
        &= \int \theta p_{\text{SBI}}(\theta \given \x_{\text{obs}}) d\theta - \int \theta \hat p_{\text{SBI}}(\theta \given \x_{\text{obs}})  d\theta \\
        &= \int \theta \left[p_{\text{SBI}}(\theta \given \x_{\text{obs}}) -  \hat p_{\text{SBI}}(\theta \given \x_{\text{obs}})\right] d\theta\\
        &= \int \theta \left[ \int p_{\text{SBI}}(\theta \given \x_{\text{obs}},\x_{\text{mis}})  p_{\text{true}}(\x_{\text{mis}} \given \x_{\text{obs}}) d \x_{\text{mis}} - \int p_{\text{SBI}}(\theta \given \x_{\text{obs}},\x_{\text{mis}}) \hat p(\x_{\text{mis}} \given \x_{\text{obs}}) d \x_{\text{mis}}  \right] d\theta\\
        &= \int \theta \int p_{\text{SBI}}(\theta \given \x_{\text{obs}},\x_{\text{mis}})\left [ p_{\text{true}}(\x_{\text{mis}} \given \x_{\text{obs}}) - \hat p(\x_{\text{mis}} \given \x_{\text{obs}}) \right] d \x_{\text{mis}} d\theta~. \\
    \end{align*}
    %
    Thus to ensure that the bias is zero, we require that $\hat p(\x_{\text{mis}} \given \x_{\text{obs}})$ be aligned with $p_{\text{true}}(\x_{\text{mis}} \given \x_{\text{obs}})$.
\end{proof}

\subsubsection{Proof for \Cref{prop:rise_objective}}
\label{subsec:proof_rise_objective}

\begin{proof}

Recall our optimization problem from \Cref{eq:alternative}:
\begin{equation*} 
          \argmin_{\psi} \,\mathbb{E}_{\mathbf{x}_{\text{obs}} \sim p_{\text{true}}} \, \text{KL}[p_{\text{true}}(\theta \mid \mathbf{x}_{\text{obs}} ) \mid \mid r_{\psi}(\theta \mid \mathbf{x}_{\text{obs}})]~.
\end{equation*}
Expanding the KL term, we note that the above is equivalent to
\begin{equation*} 
          \argmin_{\psi} \,\mathbb{E}_{\mathbf{x}_{\text{obs}} 
          \sim p_{\text{true}}} \mathbb{E}_{\theta \sim p_{\text{true}}(\theta \mid \mathbf{x}_{\text{obs}} )} \log
        \left( \dfrac{p_{\text{true}}(\theta \mid \mathbf{x}_{\text{obs}})}{r_{\psi}(\theta \mid \mathbf{x}_{\text{obs}})}\right)~.
\end{equation*}
Since $p_{\text{true}}(\theta \mid \mathbf{x}_{\text{obs}})$ does not depend on $\psi$, we immediately note that the problem is equivalent to  
\begin{align*} 
& \mathrm{argmin}_{\psi} \, \mathbb{E}_{\mathbf{x}_{\text{obs}} \sim p_{\text{true}}} \, \mathbb{E}_{p_{\text{true}}(\theta \mid \mathbf{x}_{\text{obs}} )} [- \log\, r_{\psi}(\theta \mid \mathbf{x}_{\text{obs}})]~  \\
& = \argmax_{\psi} \, \mathbb{E}_{(\mathbf{x}_{\text{obs}}, \theta) \sim p_{\text{true}}} \, [\log\, r_{\psi}(\theta \mid \mathbf{x}_{\text{obs}})]~.  
\end{align*}

We now obtain a lower bound for $\mathbb{E}_{(\mathbf{x}_{\text{obs}}, \theta) \sim p_{\text{true}}} \, [\log\, r_{\psi}(\theta \mid \mathbf{x}_{\text{obs}})]$. Formally, we have

\begin{align*}
    \mathbb{E}_{(\mathbf{x}_{\text{obs}},\theta) \sim p_{\text{true}}} [\log r_{\psi}(\theta \mid \mathbf{x}_{\text{obs}})] &= \mathbb{E}_{(\mathbf{x}_{\text{obs}},\theta) \sim p_{\text{true}}} \log \int r_{\psi}(\theta,\mathbf{x}_{\text{mis}} \mid \mathbf{x}_{\text{obs}})d\mathbf{x}_{\text{mis}} \\
    &= \mathbb{E}_{(\mathbf{x}_{\text{obs}},\theta) \sim p_{\text{true}}} \log \int \frac{p(\mathbf{x}_{\text{mis}} \mid \mathbf{x}_{\text{obs}})r_{\psi}(\theta,\mathbf{x}_{\text{mis}} \mid \mathbf{x}_{\text{obs}})}{p(\mathbf{x}_{\text{mis}} \mid \mathbf{x}_{\text{obs}})} d\mathbf{x}_{\text{mis}} \\
    &\geq \mathbb{E}_{(\mathbf{x}_{\text{obs}},\theta) \sim p_{\text{true}}} \mathbb{E}_{\mathbf{x}_{\text{mis}} \sim p(\mathbf{x}_{\text{mis}} \mid \mathbf{x}_{\text{obs}})} \left[\log  \frac{r_{\psi}(\theta,\mathbf{x}_{\text{mis}} \mid \mathbf{x}_{\text{obs}})}{p(\mathbf{x}_{\text{mis}} \mid \mathbf{x}_{\text{obs}})} \right]~\\
& = \mathbb{E}_{(\mathbf{x}_{\text{obs}},\theta) \sim p_{\text{true}}} \mathbb{E}_{\mathbf{x}_{\text{mis}} \sim p(\mathbf{x}_{\text{mis}} \mid \mathbf{x}_{\text{obs}})} \left[\log  \frac{r_{\psi}(\mathbf{x}_{\text{mis}} \mid \mathbf{x}_{\text{obs}}) r_{\psi}(\theta \mid \mathbf{x}_{\text{obs}}, \mathbf{x}_{\text{mis}})}{p(\mathbf{x}_{\text{mis}} \mid \mathbf{x}_{\text{obs}})} \right]~,
    \end{align*}
where we invoked the Jensen's inequality to swap the log and the conditional expectation. Splitting parameters $\psi$ into imputation parameters $\varphi$ and inference parameters $\phi$, and denoting the corresponding imputation and inference networks by $\hat{p}_{\varphi}$ and $q_{\phi}$ respectively, we immediately get 

\begin{align*}\mathbb{E}_{(\mathbf{x}_{\text{obs}},\theta) \sim p_{\text{true}}} [\log r_{\phi, \varphi}(\theta \mid \mathbf{x}_{\text{obs}})] \geq  \mathbb{E}_{(\mathbf{x}_{\text{obs}},\theta) \sim p_{\text{true}}} \mathbb{E}_{\mathbf{x}_{\text{mis}} \sim p(\mathbf{x}_{\text{mis}} \mid \mathbf{x}_{\text{obs}})} \left[\log  \frac{\hat{p}_{\varphi}(\mathbf{x}_{\text{mis}} \mid \mathbf{x}_{\text{obs}}) q_{\phi}(\theta \mid \mathbf{x}_{\text{obs}}, \mathbf{x}_{\text{mis}})}{p(\mathbf{x}_{\text{mis}} \mid \mathbf{x}_{\text{obs}})} \right].
\end{align*}

\noindent Thus, we obtain the following variational objective:

\begin{align*}
 & \mathrm{argmax}_{\phi, \varphi}   \mathbb{E}_{(\mathbf{x}_{\text{obs}},\theta) \sim p_{\text{true}}} \mathbb{E}_{\mathbf{x}_{\text{mis}} \sim p(\mathbf{x}_{\text{mis}} \mid \mathbf{x}_{\text{obs}})} \left[\log  \frac{\hat{p}_{\varphi}(\mathbf{x}_{\text{mis}} \mid \mathbf{x}_{\text{obs}}) q_{\phi}(\theta \mid \mathbf{x}_{\text{obs}}, \mathbf{x}_{\text{mis}})}{p(\mathbf{x}_{\text{mis}} \mid \mathbf{x}_{\text{obs}})} \right] \\
 = \,\, & \mathrm{argmax}_{\phi, \varphi}  \mathbb{E}_{(\mathbf{x}_{\text{obs}},\theta) \sim p_{\text{true}}} \left(\mathbb{E}_{\mathbf{x}_{\text{mis}} \sim p(\mathbf{x}_{\text{mis}} \mid \mathbf{x}_{\text{obs}})} \left[\log  \hat{p}_{\varphi}(\mathbf{x}_{\text{mis}} \mid \mathbf{x}_{\text{obs}}) + \log q_{\phi}(\theta \mid \mathbf{x}_{\text{obs}}, \mathbf{x}_{\text{mis}})\right] ~+~ \mathbb{H}(p(\mathbf{x}_{\text{mis}} \mid \mathbf{x}_{\text{obs}})\right) \\
 = \,\, & \mathrm{argmax}_{\phi, \varphi} \mathbb{E}_{(\mathbf{x}_{\text{obs}},\theta) \sim p_{\text{true}}} \mathbb{E}_{\mathbf{x}_{\text{mis}} \sim p(\mathbf{x}_{\text{mis}} \mid \mathbf{x}_{\text{obs}})} \left[\log  \underbrace{\hat{p}_{\varphi}(\mathbf{x}_{\text{mis}} \mid \mathbf{x}_{\text{obs}})}_{\text{imputation}} + \log \underbrace{q_{\phi}(\theta \mid \mathbf{x}_{\text{obs}}, \mathbf{x}_{\text{mis}})}_{\text{inference}}\right]~,
\end{align*}
since the entropy term $\mathbb{H}(p(\mathbf{x}_{\text{mis}} \mid \mathbf{x}_{\text{obs}})$ does not depend on the optimization variables $\phi$ and $\varphi$.\\

\color{black}

\end{proof}

\subsection{Neural process}\label{subsec:npf}
Neural Process \citep{garnelo2018conditional,foong2020meta} models the predictive distribution over target locations $\x_t$ by, (i) constructing a learnable mapping $f_\gamma$ from the context set $(\x_c,\y_c)$ to a latent representation $\mathbf{r}$ as,
\begin{align}
    \mathbf{r} = f_\gamma(\x_c,\y_c)
\end{align}
and then (ii) utilizing the representation $\mathbf{r}$ to approximate the predictive distribution, given the target locations $\x_t$, via a learnable decoder $g_\omega$ as,
\begin{align}
    p(\y_t \mid \x_t,\mathbf{r}) = g_\omega(\mathbf{r},\x_t)
\end{align}
where $\x_c,\x_t \in \mathcal{X} \subseteq \mathbb{R}^{d_x} $ are the input vectors (often locations or positions) and $\y_c,\y_t \in \mathcal{Y} \subseteq \mathbb{R}^{d_y} $  the output vectors. In practice, the predictive distribution is often assumed to factorize
as a product of Gaussians:
\begin{align}
    p(\y_t \mid \x_t,\mathbf{r}) = \prod_{m=1}^M p(\y_{t,m} \mid \x_{t,m},\mathbf{r}&) 
    = \prod_{m=1}^M \mathcal{N}(\y_{t,m} \mid \mu_{\omega,m} ,  \sigma^2_{\omega,m})
\end{align}
where $\mu_{\omega,m}, \sigma_{\omega,m} = g_\omega(\mathbf{r},\x_{t,m})$. For a fixed context $(\x_c,\y_c)$, using Kolmogorov’s extension theorem \citep{oksendal2013stochastic}, the collection of these finite dimensional distributions defines a stochastic process if these are consistent under (i) permutations of any entries of $(\x_t,\y_t)$ and (ii) marginalisations of any entries of $\y_t$.

\subsection{Implementation Details}
\label{subsec:implement}

This section is arranged as follows:
\begin{itemize}
    \item \Cref{sec:model_descriptions}: Description of SBI benchmarking simulators
    \item \Cref{subsubsec:prior}: Prior distributions used for the SBI experiments
    \item \Cref{sec:creating_mask}: Procedure for creating the missingness mask under MCAR and MNAR
    \item \Cref{sec:network_param}: Details of the neural network settings.
\end{itemize}

\subsubsection{Model descriptions}
\label{sec:model_descriptions}

\textbf{Ricker model} simulates the temporal evolution of population size in ecological systems.  In this model, the population size $N_t$ at time $t$ evolves as $N_{t+1} = N_t \exp(\theta_1)\exp(N_t + e_t), t \in \{1,\ldots,T\}$. The parameter $\exp(\theta_1)$ represents the growth rate, while $e_t$ denote independent and identically distributed Gaussian noise terms with zero mean and variance $\sigma_{e}^2$. The initial population size is set to $N_0 = 1$. Observations  $x_t$  are modeled as Poisson random variables with rate parameter $\theta_2 N_t$, such that $x_t \sim \mathrm{Poiss}(\theta_2 N_t)$.  For our simulations, we fixed $\sigma_{e}^2=0.09$ and focused on estimating the parameter vector $\theta = [ \theta_1 , \theta_2]^\top$. The prior distribution is set as a uniform distribution $\mathcal{U}([2,8] \times [0,20])$. We simulated the process for  $T=100$ time steps to generate sufficient data for inference, and considered a simulation budget of 1000 to create the dataset.

\textbf{Ornstein-Uhlenbeck process (OUP)} is a stochastic differential equation model widely used in
financial mathematics and evolutionary biology. The OU process $x_t$ is defined as,
\begin{align}
    x_{t+1} &= x_t + \Delta x_t, \quad t \in \{1,\ldots,T\}\\
    \Delta x_t &= \theta_1[\exp(\theta_2) - x_t]\Delta t+ 0.5w
\end{align}
where $T=25$, $\Delta t=0.2$, $x_0 = 10$, and $w \sim \mathcal{N}(0,\Delta t)$. 

\textbf{Generalized Linear Model (GLM).} A $10$ parameter Generalized Linear Model (GLM) with Bernoulli observations.

\textbf{Gaussian Linear Uniform (GLU).} A $10$ dimensional Gaussian model, where data points are simulated as $x \sim \mathcal{N}(x \mid \theta,\Sigma)$. The parameter $\theta$ is the mean, and the covariance $\Sigma = 0.1\mathbb{I}$ is fixed, with a uniform prior ($\theta \in \mathcal{U}([-1,1]^{10})$). We refer to \citet{lueckmann2021benchmarking,tejero-cantero2020sbi} for further details on these SBI tasks. 

\textbf{Hodgkin Huxley Model.} Hodgkin Huxley Model is a real-world computational neuroscience simulator. It describes the intricate dynamics of the generation and propagation of action potentials along neuronal membranes with the capture of the time course of membrane voltage by modeling the behavior of ion channels, particularly sodium and potassium, as well as leak currents. It consists of two parameters: $\theta_1 \equiv \bar{g}_{\text{Na}}$, and $\theta_2 \equiv \bar{g}_{\text{K}}$, which describe the density of $\text{Na}$ and $\text{K}$ specifically. The dynamics  are parameterized as a set of differential equations,
\begin{align*}
    C_{m}\frac{dV}{dt} &= g_1 (E-V) + \theta_1 m^3 h (E_{\text{Na}} - V) + \theta_2 n^4 h (E_{\text{K}} - V) + \bar{g}_{\text{M}} p (E_{\text{M}} - V) + I_{\text{inj}} + \sigma \eta(t) \\ 
    \frac{d q}{dt} &= \frac{q_{\infty}(V) - q}{\tau_q (V)}, ~~ q \in \{m,h,n,p\}
\end{align*}
Here, $V$ represents the membrane potential, $C_{m}$ the membrane capacitance, $g_1$ is the leak conductance, $E_1$ is the membrane reverse potential, $\theta_1, \theta_2$ are the densities of Na and K channel, $\bar{g}_{\text{M}}$ is the density for M channel, $E_{\text{Na,K,M}}$ denotes the reversal potential, and $ \sigma \eta(t)$ is the intrinsic neural noise. The right hand side of the voltage dynamics is composed of a leak current, a voltage-dependent $\text{Na}^+$, a delayed rectifier $\text{K}^+$,  a slow voltage-dependent $\text{K}^+$ current responsible for spike-frequency adaptation, and an injected current $I_{\text{inj}}$. Channel gating variables $q$ have dynamics fully characterized by the neuron membrane potential $V$, given the respective steady-state $q_{\infty}(V)$ and time constant $\tau_q (V)$. For more details, see \citet{pospischil2008minimal}.

\subsubsection{Prior distributions}\label{subsubsec:prior}
We utilize the following prior distributions for our experiment tasks:
\begin{itemize}
    \item Ricker: Uniform distribution $\mathcal{U}([2,8] \times [0,20])$
    \item OUP: Uniform prior $\mathcal{U}([0,2] \times [-2,2])$
    \item Hodgkin-Huxley: Uniform distribution $\mathcal{U}([10^{-4},-0.5] \times [15.0,100.0])$
    \item GLU: Uniform distribution $\mathcal{U}([-1,1]^{10})$
    \item GLM: A multivariate normal $\mathcal{N}(0,(\textbf{F}^{\top}\textbf{F})^{-1} )$ computed as follows,
    \begin{align}
        \textbf{F}_{i,i-2} = 1, \textbf{F}_{i,i-1} = -2, \textbf{F}_{i,i} = 1 + \sqrt{\frac{i-1}{9}},  \textbf{F}_{i,j} = 0 ~\text{otherwise}, 1\leq i,j \leq 9
    \end{align}

\end{itemize}

\subsubsection{Creating the missingness mask}
\label{sec:creating_mask}

\paragraph{MCAR.} We adopted random masking to simulate the MCAR scenario. For a given missingness degree $\varepsilon$, we randomly mask out $\varepsilon \%$ of the data sample.

\paragraph{MNAR.} We employed the \textit{self-masking} or \textit{self-censoring} approach as outlined by \citet{ipsen2020not}. For a given data sample $\x \in \mathbb{R}^d$, and following \citet{sinelnikov2024latent,ong2024learning}, the probability of a particular data-point to be missing depends on its value. Specifically, we sample the mask $s_i$ for $i^{th}$ value for data sample as,
\begin{align}
    s_i \sim \mathrm{Bern}(p_i), \quad p_i = {\varepsilon} \cdot \frac{x_i}{\max_d (\x)}
\end{align}
where $0 \leq i \leq d$, $\max_d (\x)$ represents the maximum value in the data sample and $p_i$ is the masking probability for data-point $x_i$ which is computed using the proportion of missing values $\varepsilon$.

\subsubsection{Network Parametrization}
\label{sec:network_param}
\paragraph{Summary Networks.} For the Ricker and Huxley model, the summary network is composed of 1D convolutional layers, whereas for the OUP, it is a combination of bidirectional long short-term memory (LSTM) recurrent modules and 1D convolutional layers. The dimension of the statistic space is set to four for both the models. We do not use summary networks for GLM and GLU.

\paragraph{Imputation Model.}
The parameters for the neural process-based imputation model used in RISE are given in \Cref{tab:hyper_ricker} and \Cref{tab:hyper_glm}.

\begin{table}[H]
\caption{Default hyperparameters for imputation model $\hat{p}_\varphi$ for Ricker, OUP and Huxley model.}
\label{tab:hyper_ricker}
\begin{center}
\resizebox{0.8\textwidth}{!}{
\begin{tabular}{lrlc}
\toprule
Module & Hyperparameter & Meaning & Value \\
\midrule
\multirow{5}{*}{Encoder} &CNN blocks & Number of CNN layers & 1 \\
&Hidden dimension & Number of output channels of each CNN layer& 64 \\
&Kernel size & Kernel size of each convolution layer & 9 \\
&Stride & Stride of each convolution layer & 1  \\
&Padding & Padding size of each convolution layer & 4  \\
\midrule
\multirow{5}{*}{Latent} &CNN blocks & Number of CNN layers & 2 \\
&Hidden dimension & Number of output channels of each CNN layer& 32 \\
&Kernel size & Kernel size of each convolution layer & 3 \\
&Stride & Stride of each convolution layer & 1  \\
&Padding & Padding size of each convolution layer & 1  \\
\midrule
\multirow{5}{*}{Decoder} &CNN blocks & Number of CNN layers & [6,1]  \\
&Hidden dimension & Number of output channels of each CNN layer& [32,2] \\
&Kernel size & Kernel size of each convolution layer & 5 \\
&Stride & Stride of each convolution layer & 1  \\
&Padding & Padding size of each convolution layer & 2  \\
\bottomrule
\end{tabular}}
\end{center}
\end{table}

\begin{table}[H]
\caption{Default hyperparameters for imputation model $\hat{p}_\varphi$ for GLM and GLU.}
\label{tab:hyper_glm}
\begin{center}
\resizebox{0.75\textwidth}{!}{
\begin{tabular}{rrlc}
\toprule
Module & Hyperparameter & Meaning & Value \\
\midrule
\multirow{2}{*}{Encoder} &MLP blocks & Number of MLP layers & [1,1] \\
&Hidden dimension & Number of output channels of each MLP layer& [32,64] \\
\midrule
\multirow{2}{*}{Latent} &MLP blocks & Number of MLP layers & 2 \\
&Hidden dimension & Number of output channels of each MLP layer& 32 \\
\midrule
\multirow{2}{*}{Decoder} &MLP blocks & Number of MLP layers & [6,1] \\
&Hidden dimension & Number of output channels of each MLP layer& [32,10] \\
\bottomrule
\end{tabular}}
\end{center}
\end{table}

\paragraph{Inference model.} Our inference model implementations are based on publicly available code from the sbi library  \url{https://github.com/mackelab/sbi}. We use the NPE-C model \citep{greenberg2019automatic} with Masked Autoregressive Flow (MAF) \citep{papamakarios2017masked} as the backbone inference network, and adopt the default configuration with 20 hidden units and 5 transforms for MAF. Throughout our experiments, we maintained a consistent batch size of 50 and a fixed learning rate of $ 5 \times 10^{-4}$.

\begin{figure}[!t]
    \centering
    \includegraphics[width=\textwidth]{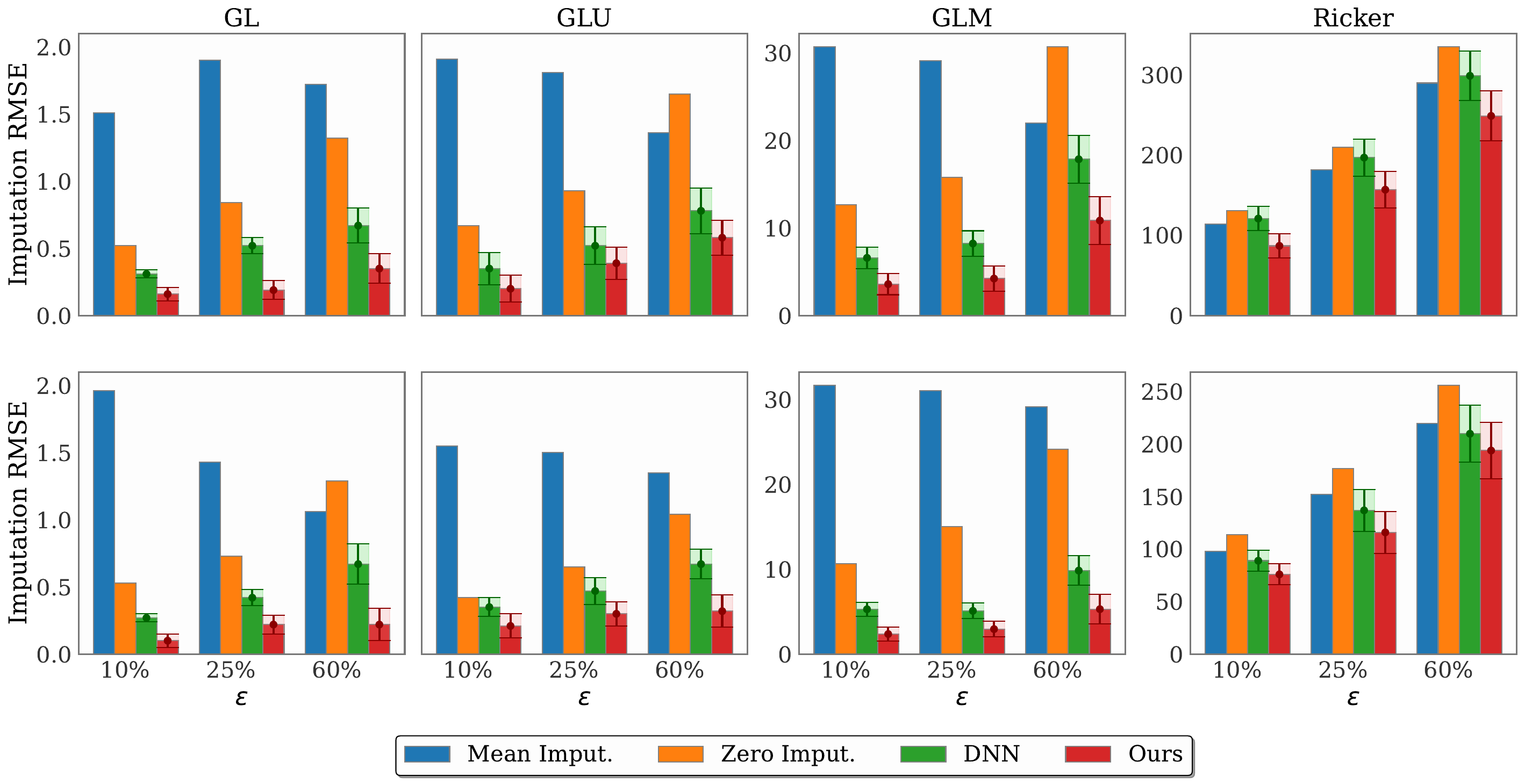}
    \caption{Imputation RMSE for MCAR (\textbf{top}) and MNAR (\textbf{bottom}) over various synthetic datasets. Here GL refers to a 10 dimension Gaussian linear model, see \citet{lueckmann2021benchmarking} for details.}
    \label{fig:imp_mse}
\end{figure}

\section{Additional Results} \label{sec:add_results}

\Cref{fig:imp_mse} shows how accurate our proposed method is in imputing the values of missing data simulated from the SBI benchmark models compared to the baselines. The performance is measured in terms of RMSE of the imputed values. Our method (denoted in red) performs the best in imputing the missing values (which eventually helps in improving the estimation of the posterior distribution).

\subsection{MMD} \label{subsec:mmd} 
In \Cref{tab:mmd}, we report the MMD values for the experiment on SBI benchmark simulators presented in \Cref{subsec:infer_syn}. Similar to the NLPP and C2ST results of \Cref{tab:nlpp_c2st}, we observe that RISE yields lowest MMD for almost all the cases, especially for Ricker and OUP where RISE beats the baselines comprehensively. 

\begin{table}[!t]
\centering
\caption{MMD under MCAR and MNAR scenarios, with missing value proportion $\varepsilon$. \ourmethod demonstrates superior posterior estimation performance. For MNAR scenarios, the proportion of missing values averages below $\varepsilon$ due to self-censoring (details in \Cref{sec:creating_mask}). Note that Simformer results are unavailable for Ricker and OUP due to the lack of official implementation.}
\resizebox{\textwidth}{!}{
\begin{tabular}{@{}llcccccccccc@{}}
\toprule
& \multirow{2}{*}{Dataset} & \multirow{2}{*}{$\epsilon$} & \multicolumn{4}{c}{MCAR}                                     &        &   \multicolumn{4}{c}{MNAR}                \\ \cmidrule{4-7} \cmidrule{9-12} 
 && & NPE-NN           & Wang et al. & Simformer        & RISE             &  & NPE-NN & Wang et al. & Simformer & RISE \\ \midrule 
\multirow{12}{*}{\rotatebox{90}{MMD}} &\multirow{3}{*}{GLU}    & $10 \%$ & $0.21 \pm 0.02$  &       $0.20 \pm 0.03$      & $0.20\pm 0.01$ & $\mathbf{0.18}\pm 0.01$  &  & $0.25 \pm 0.03$   &  $0.23 \pm 0.02$        &    $0.18 \pm 0.01$      & $\mathbf{0.16} \pm 0.01$ \\ 
 &                       & $25 \%$ & $0.27 \pm 0.02 $ &   $0.27 \pm 0.01$        & $0.27 \pm 0.01 $ & $\mathbf{0.26} \pm 0.01$ &  & $0.29 \pm 0.02 $  &  $0.26 \pm 0.02$      &  $0.25 \pm 0.01$     & $\mathbf{0.22} \pm 0.02$ \\ 
 &                       & $60 \%$ &$0.40 \pm 0.04 $ &   $0.36 \pm 0.02$         & $0.39 \pm 0.03$ & $\mathbf{0.33} \pm 0.03 $  &  &$0.33 \pm 0.02 $  &   $0.31 \pm 0.02$        &    $0.32 \pm 0.02$     & $\mathbf{0.27} \pm 0.02$ \\ 
\cmidrule{2-12}
&\multirow{3}{*}{GLM}    & $10 \%$ & $0.15 \pm 0.01$ &   $0.15 \pm 0.02$      & $0.17 \pm 0.01$ & $\mathbf{0.12} \pm 0.01 $  &  & $0.16 \pm 0.01$   &    $0.14 \pm 0.02$       &   $0.16 \pm 0.01$      & $\mathbf{0.13} \pm 0.01$ \\ 
&                        & $25 \%$ & $0.37\pm 0.02$  &  $0.30 \pm 0.02$      & $0.31 \pm 0.03$ & $\mathbf{0.27} \pm 0.03$  &  &  $0.18 \pm 0.02$ &    $0.22 \pm 0.02$         &       $0.25 \pm 0.03$   & $\mathbf{0.17} \pm 0.01$ \\  
&                        & $60 \%$& $0.50 \pm 0.04 $  &  $0.44 \pm 0.02$         & $0.52 \pm 0.03$ & $\mathbf{0.38} \pm 0.05$  &  & $0.62 \pm 0.04$   &   $0.53 \pm 0.02$        &   $0.50 \pm 0.03$   &  $\mathbf{0.47} \pm 0.02$\\  \cmidrule{2-12}
&\multirow{3}{*}{Ricker} & $10 \%$ & $0.45 \pm 0.01 $ &   $0.38 \pm 0.02$        &    -         & $\mathbf{0.31} \pm 0.01 $  &  & $0.49 \pm 0.01$   &     $0.32 \pm 0.02$     &   -        & $\mathbf{0.27} \pm 0.02$ \\ 
&                        & $25 \%$ & $0.47 \pm 0.02$ &   $0.39 \pm 0.02$         &      -        & $\mathbf{0.35} \pm 0.02$  &  &$0.49 \pm 0.02$   &    $0.41 \pm 0.02$       &  -         & $\mathbf{0.36} \pm 0.03$ \\ 
&                        & $60 \%$ &$0.51 \pm 0.02 $ &    $0.43 \pm 0.02$      &      -           & $\mathbf{0.37} \pm 0.01 $ &  & $0.57 \pm 0.01 $  &    $0.46 \pm 0.02$     &   -        & $\mathbf{0.41} \pm 0.05$ \\ \cmidrule{2-12}
&\multirow{3}{*}{OUP}    & $10 \%$ & $0.35 \pm 0.02$ &   $0.31 \pm 0.02$       &      -     & $\mathbf{0.29} \pm 0.02$ &  &$0.42 \pm 0.03$   &    $0.41 \pm 0.02$       &     -      & $\mathbf{0.38} \pm 0.03$ \\ 
 &                       & $25 \%$ &$0.36 \pm 0.02$ &   $0.33 \pm 0.02$        &    -          &  $\mathbf{0.30} \pm 0.02$&  &  $0.43 \pm 0.02$ &    $0.41 \pm 0.02$       &     -      & $\mathbf{0.38} \pm 0.02$ \\  
 &                       & $60 \%$ & $0.39 \pm 0.03$ &    $0.37 \pm 0.02$       &      -          & $\mathbf{0.35} \pm 0.02 $ & & $0.44 \pm 0.03 $   &    $0.41 \pm 0.02$      &    -    &  $\mathbf{0.39} \pm 0.02$\\ \bottomrule
\end{tabular}}

\label{tab:mmd}
\end{table}

\subsection{Coverage Plots}\label{subsec:coverage}
\begin{figure}[!hbt]
    \centering
    \includegraphics[width=0.99\linewidth]{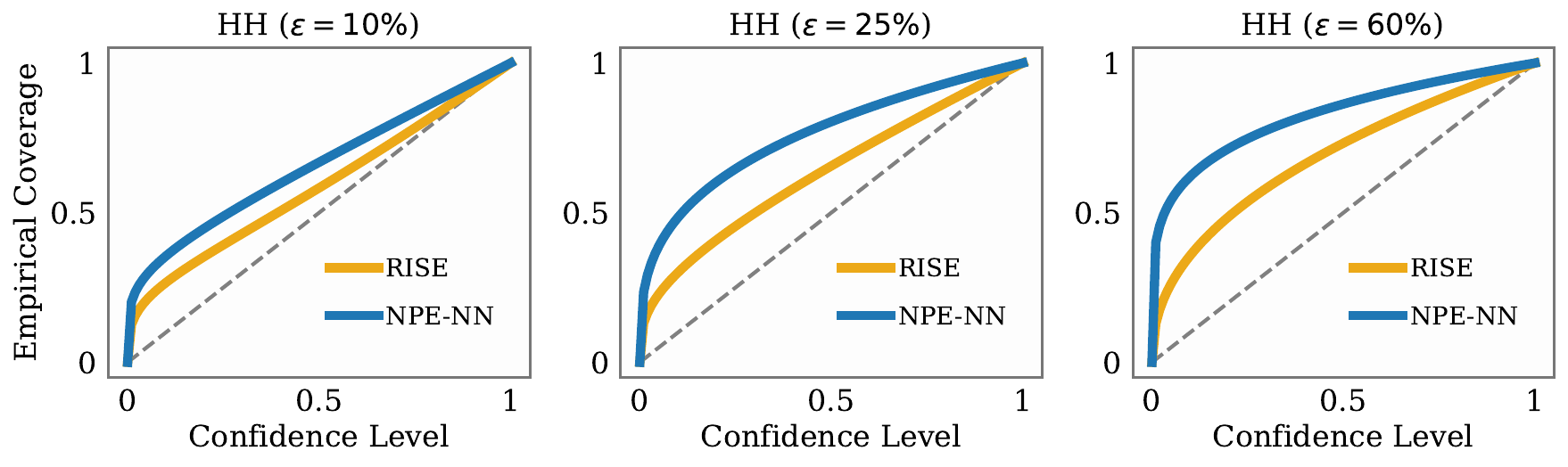}
     \includegraphics[width=0.99\linewidth]{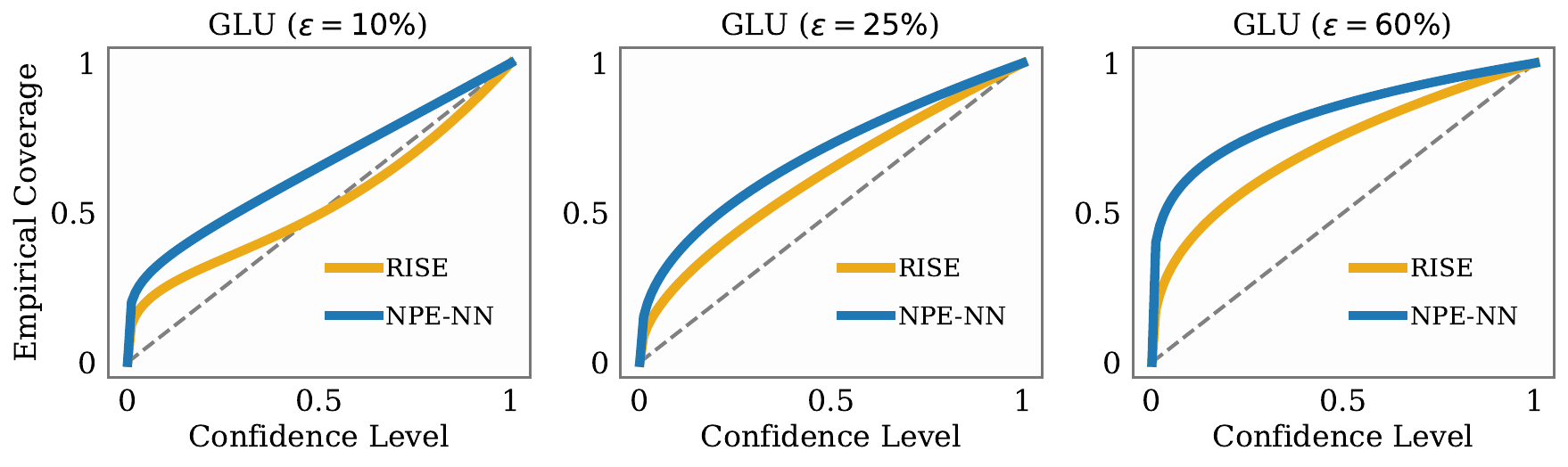}
    \caption{Expected coverage of RISE and NPE-NN for HH  (top) and GLU (bottom) task over various level of missingness. The estimator becomes more conservative with increase in missingness due to the lack of information to estimate posterior and imputation distribution. }
    \label{fig:coverage}
\end{figure}

We compute the expected coverage \citep{hermans2022a} of our method on various confidence levels. \Cref{fig:coverage} shows the expected coverage for the HH task and GLU at various levels of missingness. We observe that RISE is able to produce conservative posterior approximations, and is better calibrated than NPE-NN. 

\subsection{Additional visualizations}
\begin{figure}[!t]
    \centering
    \includegraphics[width=0.8\textwidth]{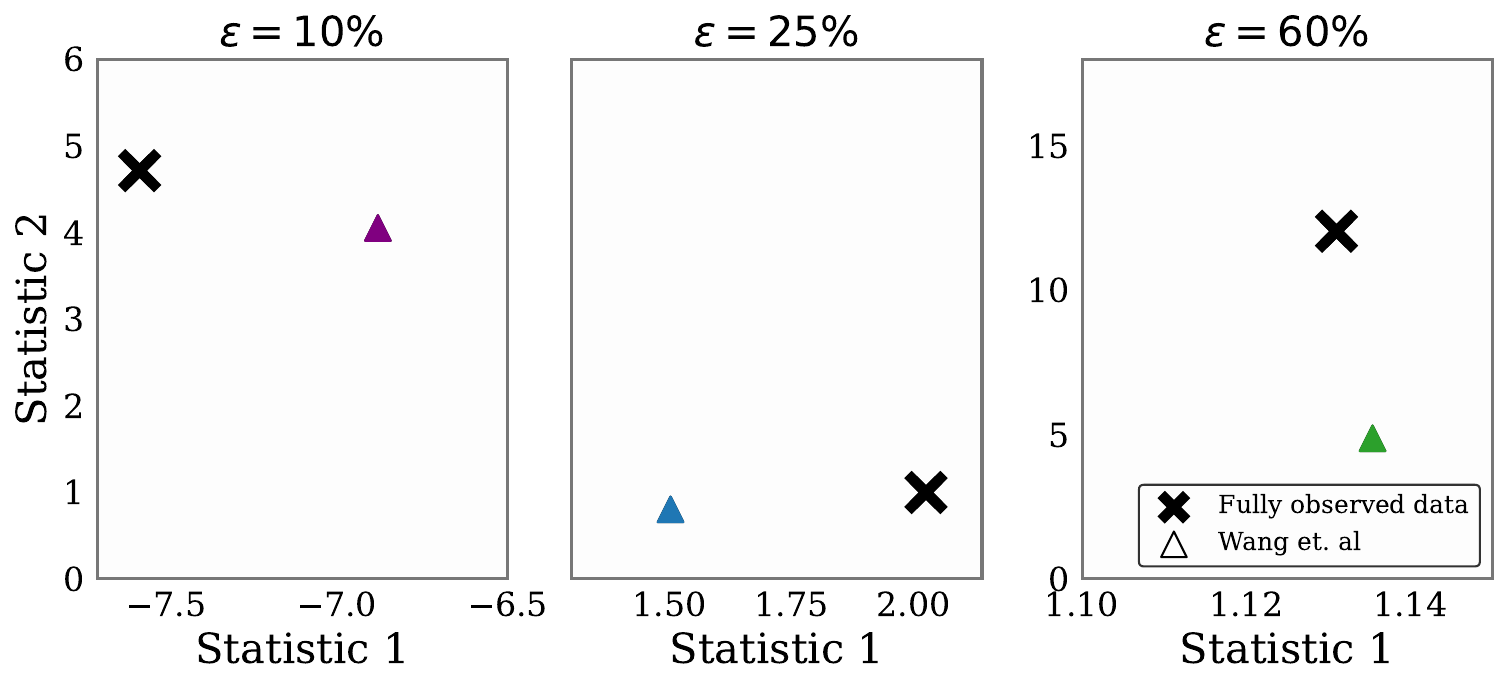}
    \caption{\textbf{Shifting Statistics.} The corresponding learned statistics for fully observed and augmented datasets as described in \cref{fig:miss_data}. Observe that the statistics for augmented datasets shift away from the fully observed statistic value, thereby leading to a shift in the corresponding NPE posterior away from the true parameter value. Note that as we re-train the method for each missingness level, the three statistics plots should not be compared with each other.}
    \label{fig:shift_stat}
\end{figure}

The \cref{fig:shift_stat} offers further insight into the posterior bias illustrated in \cref{fig:miss_data}, specifically from the perspective of learned statistics. Our observations indicate that statistics for augmented datasets deviate from the fully observed statistic value, consequently causing a shift in the corresponding NPE posterior away from the true parameter value. 

\section{Additional Ablation Studies} \label{sec:add_abl}

\begin{figure}[!hbt]
    \centering
    \subfigure[Ricker]{\includegraphics[width=0.42\textwidth]{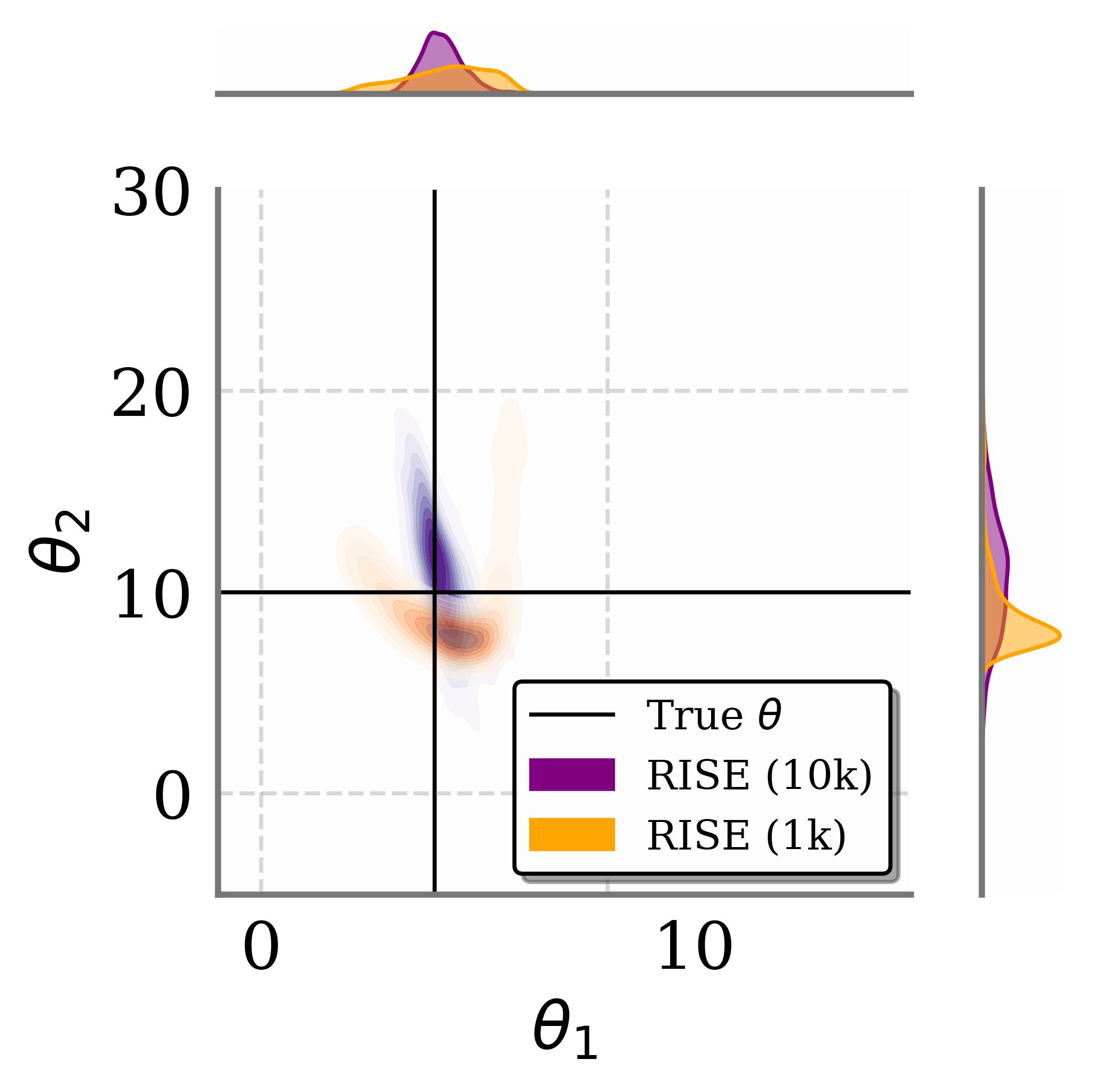}}
    \subfigure[OUP]{\includegraphics[width=0.42\textwidth]{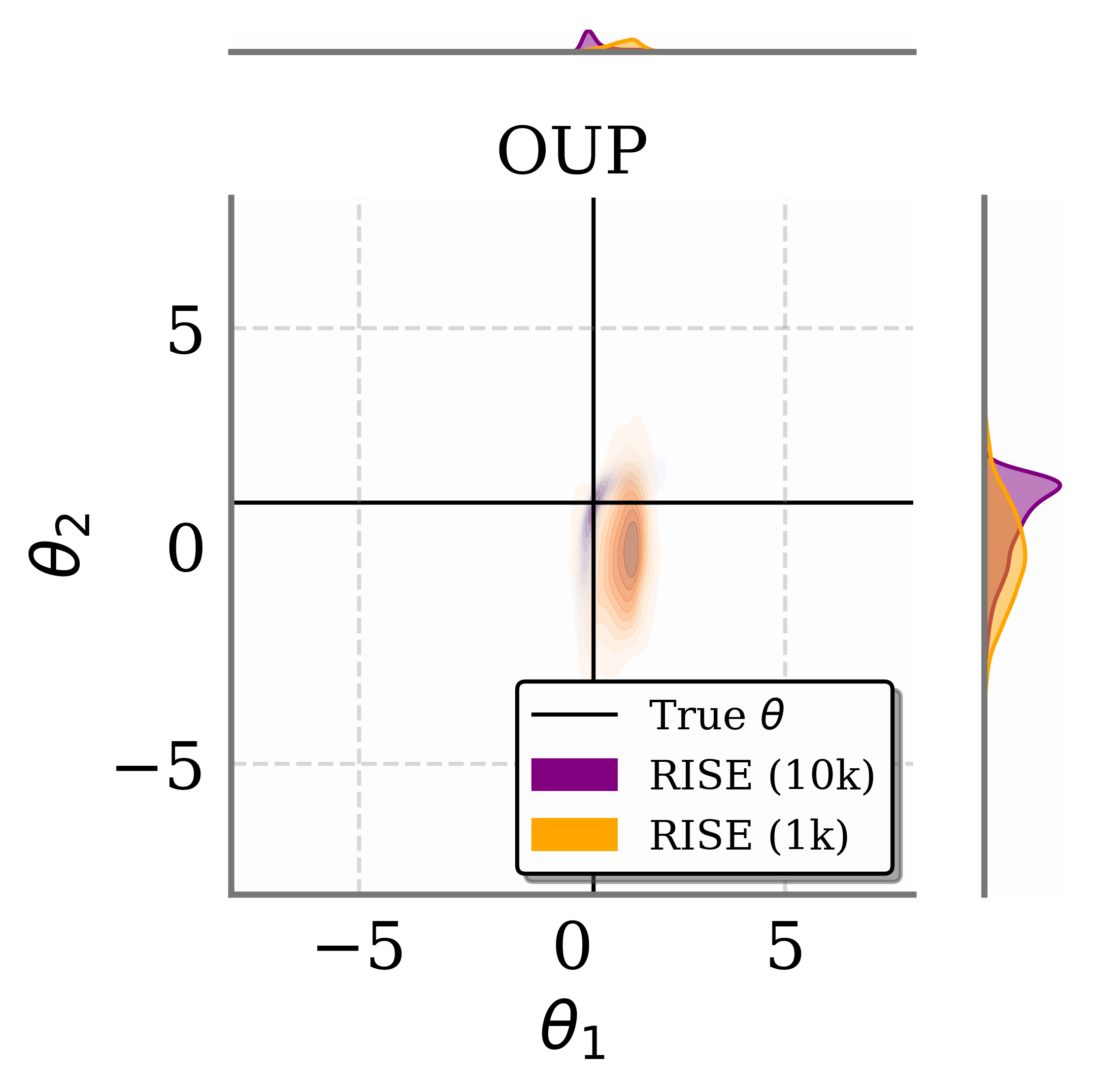}}
    \caption{Visualization of posterior estimated by RISE for Ricker and OUP under 1000 and 10000 simulation budget, respectively, and $25\%$ missingness level. We observe that the posterior estimate improves with increase in the simulation budget.}
    \label{fig:ricker_plot}
\end{figure}

\paragraph{Performance as function of simulation budget.}  We conduct a study to quantify RISE's performance as a function of the simulation budget on GLU and GLM dataset. \Cref{tab:runtime} shows C2ST and MMD for different simulation budgets for RISE, for $10 \%$ missingness level. As the budget increases, the performance improves. We also visualize the posterior obtained for different simulation budgets for Ricker and OUP in \Cref{fig:ricker_plot}.

\paragraph{Runtime comparison.} We perform an ablation study to compare the computational complexity of RISE to that of standard NPE. \Cref{tab:runtime} describes the time (in seconds) per epoch to train different models on a single V100 GPU. We observe that there is a minimal increase in runtime due to the inclusion of the imputation model. The training time remains the same with respect to missingness levels over a certain data dimensionality.

\paragraph{Flow architecture.} Our final experiment involves comparing RISE's performance for different flow architectures. We utilize neural spline flow \citep{durkan2019neural} and masked autoregressive flow as competing architectures and evaluate on the GLM model under $10\%$ missigness. \Cref{tab:abl_flow} shows that both NSF and MAF yield similar results.

\begin{table}[!hbt]
     \caption{Ablation on flow architectures  (\emph{left}) and meta learning the missingness (\emph{right}). }
    \label{tab:abl_flow}
    \vskip 0.05in
    \hspace{5pt}
    \resizebox{0.40\textwidth}{!}{
    \begin{tabular}{lccc}
        \toprule
        \textbf{Method} &  \textbf{C2ST}  &\textbf{RMSE} &\textbf{MMD} \\
        \midrule
        RISE-MAF & 0.80 & 0.65 & 0.12  \\
        RISE-NSF &  0.80 & 0.67 & 0.11\\
        \bottomrule
    \end{tabular}}
    \hspace{20pt}
    \resizebox{0.45\textwidth}{!}{
    \begin{tabular}{lc ccc}
        \toprule
        \multirow{2}{*}{Method} &  \multicolumn{2}{c}{\textbf{Ricker}}  &\multicolumn{2}{c}{\textbf{OUP}} \\
        \cmidrule{2-3} \cmidrule{4-5}
          &  RMSE  & MMD  & RMSE  & MMD \\
        \midrule
        NPE-NN  & 1.97  &  0.51 & 1.32 &  0.50 \\
        RISE-Meta  & \textbf{1.52}  & \textbf{0.42} & \textbf{0.89} & \textbf{0.45}\\
        \bottomrule
    \end{tabular}}
    
\end{table}

\begin{table}[!hbt]
     \caption{Runtime comparisons (\emph{left}) and Simulation budget comparisons (\emph{right}). }
    \label{tab:runtime}
    \vskip 0.05in
    \hspace{35pt}
    \resizebox{0.30\textwidth}{!}{
    \begin{tabular}{llc}
        \toprule
        \textbf{Method} &  \textbf{GLM}  &\textbf{GLU} \\
        \midrule
        NPE & 0.12 &  0.10\\
        RISE & 0.18 & 0.16 \\
        \bottomrule
    \end{tabular}}
    \hspace{30pt}
    \resizebox{0.45\textwidth}{!}{
    \begin{tabular}{ll cc cc}
        \toprule
        \multirow{2}{*}{}{Budget} &  \multicolumn{2}{c}{GLU}  &\multicolumn{2}{c}{GLM}  \\
        \cmidrule{2-5}
         & C2ST & MMD &C2ST  &MMD  \\
        \midrule
        1000 & 0.83 & 0.18 & 0.80&0.12  \\
        10000 & 0.78 & 0.15 & 0.75 & 0.10 \\
        \bottomrule
    \end{tabular}}
    
\end{table}

%




\color{black}
\end{document}